\documentclass[twoside]{article}

\usepackage[accepted]{aistats2025}

\usepackage[toc,page,header]{appendix}
\usepackage{minitoc}

\doparttoc %
\faketableofcontents %

\usepackage[round]{natbib}

\usepackage{url}            %
\usepackage{booktabs}       %
\usepackage{amsfonts}       %
\usepackage{nicefrac}       %
\usepackage{microtype}      %
\usepackage{xcolor}         %
\usepackage{amsmath}
\usepackage{enumitem}

\usepackage{amsfonts}       %
\usepackage{graphicx}
\usepackage{amssymb}
\usepackage{amsthm}
\usepackage{bm}
\usepackage{array}          %
\usepackage{bbm}
\usepackage{booktabs}
\usepackage{multirow}

\usepackage{wrapfig}

\usepackage{amsmath,amsfonts,bm}

\def\eqref#1{equation~\ref{#1}}

\def\1{\bm{1}}

\DeclareMathAlphabet{\mathsfit}{\encodingdefault}{\sfdefault}{m}{sl}
\SetMathAlphabet{\mathsfit}{bold}{\encodingdefault}{\sfdefault}{bx}{n}

\newtheorem{prop}{Proposition}

\definecolor{mydarkblue}{rgb}{0,0.08,0.45}
\usepackage[colorlinks=true,
    linkcolor=mydarkblue,
    citecolor=mydarkblue,
    filecolor=mydarkblue,
    urlcolor=mydarkblue,
    pdfview=FitH]{hyperref}
\usepackage{cleveref}
    
\renewcommand{\vec}[1]{\mathbf{#1}}
\renewcommand{\Re}{\mathbb{R}}
\newcommand{\eqdef}{\,{\buildrel \text{def} \over =}\,}
\newcommand{\W}[2]{\vec{W}^{(#1)}_{#2}}
\newcommand{\h}[1]{\vec{h}^{#1}}
\newcommand{\f}[1]{\vec{f}^{#1}}

\usepackage{algorithm}
\usepackage{algorithmicx}
\usepackage{algpseudocode}

\algnewcommand\algorithmicpredict{\textbf{Predict:}}
\algnewcommand\Predict{\item[\algorithmicpredict]}
\renewcommand{\vec}[1]{\mathbf{#1}}
\renewcommand{\Re}{\mathbb{R}}

\DeclareMathOperator{\loss}{\mathcal{L}}

\newcommand{\process}{\mathcal{T}}
\newcommand{\dataset}{\mathcal{D}}
\newcommand{\mask}{\mathbf{O}}
\newcommand{\maskij}{\mathbf{o}}

\newcommand\blfootnote[1]{%
  \begingroup
  \renewcommand\thefootnote{}\footnote{#1}%
  \addtocounter{footnote}{-1}%
  \endgroup
}

\begin{document}

\twocolumn[

\aistatstitle{MODL: Multilearner Online Deep Learning}

\aistatsauthor{ Antonios Valkanas$^{\dagger}$ \And Boris N. Oreshkin$^{\ddagger}$ \And  Mark Coates }

\aistatsaddress{ McGill University, Mila$^{\star}$, ILLS$^{\star}$ \And  Amazon Science \And McGill University, Mila$^{\star}$, ILLS$^{\star}$ } ]

\begin{abstract}
Online deep learning tackles the challenge of learning from data streams by balancing two competing goals: fast learning and deep learning. However, existing research primarily emphasizes deep learning solutions, which are more adept at handling the ``deep'' aspect than the ``fast'' aspect of online learning. In this work, we introduce an alternative paradigm through a hybrid multilearner approach. We begin by developing a fast online logistic regression learner, which operates without relying on backpropagation. It leverages closed-form recursive updates of model parameters, efficiently addressing the fast learning component of the online learning challenge. This approach is further integrated with a cascaded multilearner design, where shallow and deep learners are co-trained in a cooperative, synergistic manner to solve the online learning problem. We demonstrate that this approach achieves state-of-the-art performance on standard online learning datasets. We make our code available: \url{https://github.com/AntonValk/MODL}
\end{abstract}

\section{INTRODUCTION}
\blfootnote{$\dagger$ Contact: \texttt{antonios.valkanas@mail.mcgill.ca}}
\blfootnote{$\ddagger$ This work does not relate to the author's position at Amazon.}
\blfootnote{$\star$ ILLS: International Laboratory on Learning Systems, Mila - Quebec AI Institute}
\setcounter{footnote}{0}
Off-line machine learning algorithms are trained on bulk datasets, making multiple passes over them to gradually tune model parameters. In numerous scenarios, it is advantageous to learn from data streams by processing each instance sequentially, giving rise to a method known as \emph{Online Learning}~\citep{EonBottou1998OnlineLA}. Compared to batch training, online learning enables scalable and memory-efficient training, even with unbounded dataset sizes. Research in online learning spans several decades and encompasses supervised~\citep{zinkevich2003OGD}, semi-supervised~\citep{belkin2006unlabeled}, and unsupervised learning~\citep{guha2003clustering} archetypes.
Early methods concentrated on shallow learners, which offer fast learning but limited expressive capacity~\citep{hoi2013kernel}. In contrast, deep neural networks provide extensive expressive power, though their learning process is typically much slower.
\citet{sahoo2018ODL} tackle this issue via the ODL (Online Deep Learning) framework, which simultaneously learns both neural network parameters and architecture ``hedge'' weights. ODL is based on \emph{Hedge Backpropagation}~\citep{freund1997backprop}, which alternates between standard backpropagation and a hedge optimization step that adjusts the scalar weights connecting early exit points of the deep neural network. More recently,~\citet{agarwal2023auxdrop} extended ODL to address the problem of unreliable features in data streams. While these methods partially mitigate the challenges of online neural network training, they continue to face computational complexity and stability issues due to their dependence on hedge backpropagation. At a higher level of abstraction, hedge backpropagation can be viewed as a joint architecture learning task with two interdependent optimization objectives—hedge weights and neural network weights—which can interfere with one another, ultimately slowing the learning process.

\textbf{Contributions}. To overcome these inefficiencies, we present the Multilearner Online Deep Learning (MODL) framework. In contrast to prior approaches, MODL removes hedge backpropagation, a common feature in recent state-of-the-art architectures, and advances the field by: (i) introducing a novel, faster overall framework for jointly learning neural network parameters and topology, and (ii) employing efficient statistical approximations to accelerate learning in a subset of learners via closed-form recursive updates. 
In summary, our contributions are threefold: 
\vspace{-0.1cm}
\begin{itemize}[leftmargin=*]
\itemsep 0em
\item We propose MODL, a novel framework for online deep learning that employs a stacking architecture and hybridizes backpropagation with closed-form updates to accelerate the online training process;
\item We derive a fast recursive logistic regression algorithm capable of learning from data streams;
\item MODL attains state-of-the-art convergence speed and accuracy on standard benchmarks.

\end{itemize}

\section{RELATED WORK}

\textbf{Multiple learners.} Using multiple learners is a well-established idea in machine learning. Two key advantages of using multiple learners are improved point prediction and uncertainty quantification~\citep{lakshminarayanan2017ensembles}. 
Ensemble methods rely on the predictions of multiple models (ensemble members) to produce an overall prediction. Perhaps the most simple way to aggregate the model predictions is a straightforward model average~\citep{dietterich2000ensemble}.
Similarly, ensembles can provide uncertainty estimates by calculating the variance of model predictions to provide a confidence interval. The idea of using many models to prevent overfitting, such as bagging for random forests, has been well known for decades~\citep{hastie2009elements}. 
Our online learning problem setting does not suit standard ensemble methods that combine the predictions of independently trained base learners trained with the full batch. While there are strategies such as~\citep{shui2018diversity, dangelo2021repulsive, masegosa2020misspecification} for jointly training the constituent members of an ensemble, they do not address online learning and are not trivial to adapt. 

Another related area is model cascading~\citep{weiss2010cascades}. Most standard cascading models do not address online learning~\citep{chen2012classifier}. A notable work on cascade learning that works online is from~\citet{nie2024cascade}. This work views model cascading as a way to select the minimally expensive model, from a set of large language models, that can answer a natural language query. However, the aim of the work is learning to select the minimal cost model and using that model alone to predict the output. Our work focuses on using the predictions of all models and learning how to combine them effectively with the proposed architecture.

\textbf{Online learning.} The field of online learning has evolved significantly over the past three decades, with early statistical methods tracing back to foundational work by \citet{EonBottou1998OnlineLA}. A pivotal finding by \citet{bottou2003online} demonstrated that online learning algorithms can achieve learning efficiency superior to traditional full batch methods. However, this result is hard to achieve in practice. This revelation laid the groundwork for the widespread adoption of first-order algorithms that aim to learn efficiently via gradient descent. These methods are favored due to their simplicity and low cost~\citep{zinkevich2003OGD, bartlett2007adaptive}. 
A drawback of gradient approaches is that it can be tricky to select the correct learning rate for the algorithm, since there is no validation data available, due to the online nature of the learning task. 

Algorithms that employ filtering style updates, such as online generalized linear models (GLMs), can set their own learning rates. 
One such GLM is online logistic regression, and this has been explored in recent efforts that focused on online optimization. \citet{agarwal2021folklore} proposed an iterative optimization scheme that regrettably lacks the closed form updates that are needed to improve efficiency. \citet{vilmarest2021kalman} use an extended Kalman filter framework. In this paper we derive online logistic regression directly from the Bayesian approximation of the posterior and also extend the online filtering results to multinomial logistic regression. Our online logistic regression utilizes inherently fast first-order updates. 

While second-order methods have been explored in online learning~\citep{hazan2007convex,drezde2008confidenceWeighted} and can accelerate parameter convergence, they often come with significant computational overhead due to costly Hessian calculations. To reconcile the speed of the first-order methods with the sample efficiency of the second-order methods, \citet{sahoo2018ODL} proposed the hedge-backpropagation update. This approach enables simultaneous optimization of both the neural network architecture and parameters, providing a mechanism for dynamic online adjustment of model depth. Building on this work, \citet{agarwal2023auxdrop} proposed a scalable method that addresses another key challenge in streaming data: feature reliability. Previously, the issue of unreliable features had only been tackled within the scope of traditional, non-deep learning approaches~\citep{beyazit2019OLVF,he2019capricious}. While~\citet{agarwal2023auxdrop} propose a first step to handling haphazard feature, they still assume that some base features are always available.

Our approach moves the field forward by departing from hedge backpropagation updates, which have formed the basis of state-of-the-art approaches for the past few years. In this paper, we argue that the use of hedge backpropagation is suboptimal from a learning speed and performance perspective. Instead, we propose a new direction based on a stacking framework that co-trains multiple models. To handle unreliable features we propose a set learner that accepts a variable number of inputs. This eliminates the need for deterministic dropout~\citep{agarwal2023auxdrop} and streamlines the gradient flow. An additional, advantage of our set learner approach is that we eliminate the assumption of always available base features.

\textbf{Motivation \& Applications} 

Online learning has three main application areas. First, it is useful in situations where retaining training data is undesirable due to privacy concerns or legal constraints. To reduce the risk of data breaches, sensitive information can be utilized for training without being stored~\citep{yang2022burn}, necessitating the ability to train models with a single pass over the data~\citep{min2022onePass}. Some data are ephemeral by law. Regulations often mandate that certain financial transaction records (\textit{e.g.}, from online purchases) be retained for only a short duration to protect consumer privacy. Under laws like HIPAA in the United States, patient data must be handled with strict confidentiality, and certain data must be purged or anonymized after specific timeframes, especially for non-essential records. 

As a result, training typically begins with a limited dataset, followed by online learning as new data become available~\citep{Graas2023JustintimeDL}. 
Second, in data-intensive environments such as IoT devices~\citep{omar2022iot}, real-time surveillance systems~\citep{zhang2020videoStream}, or recommendation systems~\citep{valkanas2024personalized}, storing and training on the entire data stream is impractical, making online learning essential. For example, the enormous volumes of data generated by experiments at CERN\footnote{\url{https://home.cern/science/computing/storage}} are largely discarded after short storage periods, with only minimal amounts retained~\citep{Basaglia2023DataPI}. 
Third, online learning is crucial for adapting models to discrepancies between training and deployment environments, as well as to handle distribution shifts and domain generalization---an important challenge in healthcare machine learning~\citep{ktena2024generative}. For example, adaptation in medical image segmentation~\citep{valanarasu2022onthefly, ray2024adolescent, thapaliya2025gnn, suresh2023imaging, sapkota2024multimodal, thapaliya2025dsam} and credit card~\citep{mienye2024credit} or insurance fraud detection~\citep{zhang2024pre} frequently requires online training to adjust for the mismatch between training and inference distributions.

\section{PROBLEM STATEMENT}

We address the Online Learning task with data streams that contain missing features. In this problem setting the data generating process produces a sequence of triplets $\process=\{(\vec{z}_1, \vec{y}_1, \mask_1),\dots, (\vec{z}_T, \vec{y}_T, \mask_T)\}$ sampled sequentially in $T$ time steps. $\process$ consists of input features $\vec{z}_t \in \mathbb{R}^d$, ground truth labels $\vec{y}_t \in \mathcal{Y}$ (where $\mathcal{Y}$ has a fixed dimension and may be discrete or continuous depending on the task) and the set of available input indices $\mask_t =(\maskij_{j})_t \in \{0,1\}^{d}$. $\mask_t$ is a binary mask encoding observed entries, \emph{i.e.}, it is a vector of missingness indicators such that $(\maskij_{j})_t=1$ if $\vec{z}_t(j)$ is observed, and $(\maskij_{j})_t=0$ otherwise. During training we do not have access to $\process$; we can only observe an incomplete dataset $\dataset$. Denoting ``not available'' by NA, we have:
\begin{align}\label{eq:problem_setup}
    \vec{x}_t &= \vec{z}_t \odot \mask_t + {\tt NA}\odot (\mathbf{1}_{d}-\mask_t), \\
    \dataset&=\left[(\vec{x}_1, \vec{y}_1), \dots, (\vec{x}_T, \vec{y}_T)\right].
\end{align}
Since the problem setting is online learning, the goal is to train a new model from scratch in a streaming dataset setup such that at  time $t$, we have access to only the input features $\vec{x}_t$,the mask $\mask_t$, and the previous model. The input to the model is a concatenated vector that consists of $\vec{x}_t$ and $\mask_t$ and has length $2d$. After a prediction $\widehat{\vec{y}}_t$ is made, we obtain the output labels $\vec{y}_t$. We do not have access to any previous training examples so each datapoint is used for training only once. Online learning models are evaluated by the cumulative predictive error across all time steps $E_\textsc{total}=\sum_t \epsilon(\vec{y}_t, \widehat{\vec{y}}_t)$. In general, $\epsilon(\cdot,\cdot)$ is a non-negative valued cost function.

\section{METHODOLOGY}

The main drawback of neural networks in the online setting is their sensitivity to hyperparameters. For example, it can be very difficult to know the optimal width or learning rate of the network before the dataset has been streamed. Deep neural networks require low learning rates to be stable during training, but this entails slow learning 
and accumulating large errors for millions of optimization steps. On the other hand, fast learners are typically shallow and train quicker, but lack the ability to learn complex representations. To address the hyperparameter issue as well as learner complexity and learning speed trade-off, we propose a new foundational online learning architecture called MODL: Multilearner Online Deep Learning; this is the main contribution of this paper. This architecture compensates for the limitations of neural learners in the online setting by employing multiple learners along different points of the complexity and learning speed trade-off and with different hyperparameters. Our architecture then automatically selects the best weighted combination of these learner's outputs to minimize the distributional divergence between the predicted distribution and the true data distribution.

\subsection{Multilearner Online Deep Learning}

Consider a probability space $(\Omega, \mathcal{A}, \mathcal{P})$ with sample space $\Omega$, sigma algebra $\mathcal{A}$ of subsets of $\Omega$, and a convex class probability measure $\mathcal{P}$ on $\Omega$. We define a scoring function $S: \mathcal{P} \times \Omega \mapsto \overline{\Re}$, where $\overline{\Re}$ is the extended real line. We only score forecasts $P\in\mathcal{P}$ that are integrable for $\omega\in\Omega$, denoting the score as $S(P, \omega)$. To compare two probabilistic forecasts $P, Q$, we compute $S(P,Q)=\int S(P, \omega)\, dQ(\omega)$. Scoring rules induce divergence metrics $d(P,Q) : \mathcal{P} \times \mathcal{P} \mapsto (0, \infty) \eqdef S(Q,Q)-S(P,Q)$. A popular score in the literature that we also adopt in this paper is the logarithmic score, yielding $d(p,q)=\text{KL}(q,p)$, where KL is the Kullback-Leibler divergence~\citep{yao2018using, yao2024stacking}.
\begin{figure*}[t]
    \centering
    \includegraphics[]{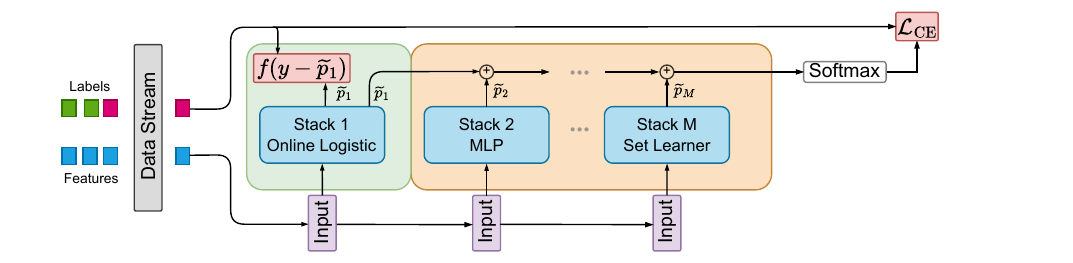}
    \caption{Multilearner Online Deep Learning. The dataset is streamed sequentially. Fast learners quickly adapt to the data distribution, providing a strong baseline for the deeper models. By synergizing models with different bias-variance trade-offs, the overall architecture quickly adapts to the data and learns deep representations. Individual model latent class scores $\widetilde{p}_i$ are sum pooled, then projected into class probabilities. During co-learning models learn to predict on top of each other. Non-neural learners (green box) learn via filtering style updates; $f(y-\widetilde{p}_1)$ is \cref{eq:update2}. Neural learners (orange box) learn via backpropagation of cross entropy loss $\loss_{\text{CE}}$.}
    \label{fig:model}
\end{figure*}
Having defined the score, we may now define the overall stacking objective as an optimization. Consider the collection of $K$ candidate models $\mathcal{M}=\{ \mathcal{M}_1, \mathcal{M}_2, \dots, \mathcal{M}_K\}$ that produce $K$ predictive distributions $\mathcal{P}=\{ \mathcal{P}_1, \mathcal{P}_2, \dots, \mathcal{P}_K\}$. 
For each model we generate a prediction $\widehat{p}(\vec y_t|\vec x_t, \mathcal{M}_i, \theta_i)$, where $\theta_i$ are the model parameters, and assign weight $w_i$ to this prediction. The predicted density is
\begin{align}
    \widehat{p}(\vec y_t|\vec x_t) = \sum_{i=1}^K w_i \, \widehat{p}(\vec y_t|\vec x_t, \mathcal{M}_i, \theta_i).
\end{align}
We may obtain the stacking weights by following optimization:
\begin{align}
    &\min_{w} \quad  d_\text{KL}\left(\sum_{i=1}^K w_i \, \widehat{p}(\vec y_t|\vec x_t, \mathcal{M}_i, \theta_i), p(\vec y_t|\vec x_t)\right)\\
    &\text{subject to }  \sum_{i=1}^K w_i = 1, \quad w_i \geq 0,
\end{align}
where $p(\vec y_t|\vec x_t)$ is the true distribution.
In practice, we  optimize over an empirical approximation by sampling the distribution for $T$ time steps yielding the objective:
\begin{align}\label{eq:stack_opt}
    \min_{w} \quad  \sum_{t=1}^T d_\text{KL}\left(\sum_{i=1}^K w_i \, \widehat{p}(\vec y_t|\vec x_t, \mathcal{M}_i, \theta_i), p(\vec y_t|\vec x_t)\right).
\end{align}
Since it is standard to use negative log-likelihood (NLL) to train the neural parameters we jointly optimize over $w_i$ as well as the parameters $\theta_i$. We modify the previous optimization due to equivalence of minimizing KL divergence and NLL:
\begin{align}\label{eq:nll}
    \min_{w, \theta} -\sum_{t=1}^T \log \left(\sum_{i=1}^K w_i\widehat{p}(\vec y_t|\vec x_t, \mathcal{M}_i, \theta_i)\right).
\end{align}
We interpret $\widetilde{p}_i=w_i\widehat{p}(\vec y_t|\vec x_t, \mathcal{M}_i, \theta_i)$ as the latent class score per model $i$. The latent class scores can be interpreted as an unnormalized/improper probability measure. Note that each neural model directly outputs the weighted prediction $\widetilde{p}_i$ which is not a valid probability distribution in general. 
The optimization in~\cref{eq:nll} requires $w_i$ to sum to 1. Since this is not the case as the neural network output is only constrained to be positive we apply softmax to the sum to project to a valid probability distribution as shown in~\Cref{fig:model}.
This architectural decision differentiates our approach from ensembling as we do not aggregate predicted class probabilities and we do not train models independently of each other. 
Instead, the latent class scores are aggregated via sum pooling and then passed through one softmax layer to produce the weighted final prediction. Thus the $i$-th model $f_i$,  $f_i(\vec x_t, \theta_i) \equiv w_i\widehat{p}(\vec y_t|\vec x_t, \mathcal{M}_i, \theta_i)$, learns both the associated probability distribution and its weight, implicitly, while the final prediction has the following form:
\begin{align}
\widehat{p}(\vec y_t|\vec x_t) = \textrm{softmax}\left[\sum_{i=1}^K f_i(\vec x_t, \theta_i)\right].
\end{align}
We demonstrate in our experiments that the proposed modeling approach enables mixing of the learners with heterogeneous learning paradigms (e.g. backpropagation and approximate closed-form) and is more sample effective compared to the procedures that attempt learning both weights and architectures explicitly (e.g. hedge backpropagation based Aux-Drop or ODL). Since the scores $f_i(\cdot)$ are not probabilities, they are unbounded (besides being non-negative), therefore, models that are confident in their prediction assign larger scores for a particular class. This is an implicit and learnable way for each model to select its own confidence level.

\textbf{Selecting the learners} The learners, which can be interpreted as functional priors, must be selected by the user before training. One could argue that this is a limitation as depending on the group of learners we select MODL can perform differently.
We, however, argue that this is a good thing as it forces the user to explicitly state their assumptions. By selecting only deep models the user assumes that the distribution is too difficult to learn using a linear learner. Conversely, by selecting many models along various points of the bias-variance trade off, the user demonstrates ignorance of the complexity of the joint feature-label distribution. 
In practice, to select the group of learners we propose the following workflow: (i) select powerful generic architectures that are applicable to the problem at hand; (ii) define the \textit{smallest} and the \textit{largest} capacity models that would be plausibly expected to learn the data distribution; (iii) define several models in-between the largest and the smallest capacity models; (iv) train the aforementioned models under the MODL framework.

In the rest of the methodology section, we propose a fast learner and a slow but powerful set learner, along with a standard MLP that serves as a medium learner. We employ these learners in all our experiments with universally good results.
Here, the fast learner based on online logistic regression quickly learns a linear approximation to the solution and provides the strongest signal guiding the global model output initially. As more data arrive, the second stage learner, implemented as an MLP, kicks in, becoming the most accurate learner before the heavier model has had the time to converge sufficiently.
Finally, the set learner, which has a deep structure and the ability to represent the semantics of variables, provides the modeling output that is the finest and therefore the hardest to learn. For example, learning semantic embeddings of variable IDs may take a long time, but this endows this part of the architecture with the ability to handle missing data and encode complex input-output relationships. Of course, this can be generalized beyond the three aforementioned learning levels. We choose to outline the structure used in our experiments, which also happens to capture the most important methodological thinking of our approach.

\textbf{Fast Learner}. Consider the standard task of fitting a logistic regression model. Denote the dataset by $\dataset = \{(\vec x_i, y_i)\}_{i=1}^n$, with $\vec x\in\Re^{m}, y_i\in\{0,1\}$. For a simple generalized linear model $p(y|\vec x, \theta) = \sigma(\vec x \theta)$, with weights $\theta\in\Re^m$, where $\sigma(\cdot)$ is the logistic function, we assign a normal prior $p(\theta) = \mathcal{N}(\theta; \vec m_0, \vec P_0)$, with mean $\vec m_0\in\Re^m$ and symmetric positive definite covariance matrix $\vec P_0\in\Re^{m\times m}$. Proposition~\ref{prop:glm} derives a closed form filtering style update for the model parameters for each observation.
\begin{prop}\label{prop:glm}
    Assuming an input feature distribution for $\vec x$ that is approximately normal, and linearizing the non-linear relationship, $\vec y = \sigma(\theta \vec x)$, 
    a quadratic approximation to the posterior of model weights after observing the $n$-th datapoint is given by the recursive formula $p(\theta | \{(\vec x_i, y_i)\}_{i=1}^n) \approx \mathcal{N}(\theta; \vec m_n, \vec P_n)$. For logistic regression we take $\sigma(\cdot)$ to be the logistic function.
\end{prop}
The proof, concrete formulae yielding $\vec m_n, \vec P_n$, and multinomial extension are provided in~\Cref{app:derivation}.

\begin{algorithm}[t]
	\caption{Online Bayesian Logistic Regression}\label{alg:logistic}
\begin{algorithmic}
  \Require Dataset $\dataset = \{(\vec x_i, y_i)\}_{i=1}^n$\\
  \textbf{Initialize}: $p(\theta)=\mathcal{N}(\theta; \vec m_0, \vec P_0)$, $\vec m_0 = 0$, where $\mathbf{P_0} = \mathbf{I}$\\
  \textbf{Return} $p(\theta | \{(\vec x_i, y_i)\}_{i=1}^n) = \mathcal{N}(\theta; \vec m_n, \vec P_n)$
 \For{t = 1,\dots,n}
		\State Receive instance: $\vec{x}_t$ and predict: $\hat{y}_t = \sigma(\vec x_t\vec m_{t-1})$
		\State Reveal true value: $y_t$ and update parameters: 
          \begin{align}
            \Xi_t &= \vec x_t \vec P_{t-1} \vec x_t^\top + \vec P \left[\left(1-\vec \sigma(\vec x_t \vec m_{t-1})\right) \sigma\left(\vec x_t \vec m_{t-1}\right)\right]^2,\\
            \vec K_t &=  \vec P_{t-1} \vec X_t^\top \left(1-\vec \sigma(\vec x_t \vec m_{t-1})\right) \sigma\left(\vec x_t \vec m_{t-1}\right)\Xi_t^{-1},\\
            \vec m_t &= \vec m_{t-1} + \vec K_t [\vec y_t - \sigma\left(\vec x_t \vec m_{t-1}\right)],\label{eq:update2}\\
            \vec P_{t} &=  \vec P_{t-1} - \vec K_t \Xi_t \vec K_t^\top.
        \end{align}
		\EndFor
	\end{algorithmic}
\end{algorithm}
As a consequence of this recursive formula, we derive the sequential~\Cref{alg:logistic}, based on the quadratic approximation to the log-likelihood. This algorithm can process the dataset in one pass, while never storing any data in memory.
Note that $\Xi_t^{-1}$ in~\Cref{alg:logistic} is a scalar, so there is no need to invert any matrices in our approach. The only memory requirement is storage of the parameters of the normal posterior distribution. 
The parameters update proportionally to the ``innovation'' (\cref{eq:update2}), which is data dependent, rather than set by the user. This fast learner is a high bias and low variance model. Next, we propose a highly expressive module that lies at the opposite end of the bias-variance spectrum.

\textbf{Slow learner with set inputs.} Rather than masking missing features with a zero or using deterministic dropout, as is done in~\citep{agarwal2023auxdrop}, we treat the input as a set that excludes any missing features.
Recall that the data generating process produces a sequence of triplets $\process=\{(\vec{z}_1, \vec{y}_1, \mask_1),\dots, (\vec{z}_T, \vec{y}_T, \mask_T)\}$. 
Then, the set of input features $\mathcal{X}_t$ can be expressed as:
\begin{align}%
    \mathcal{X}_t &= \left\{\vec{z}_{t,j} : \mask_{t,j} = 1\right\}, \quad 
    \mathcal{I}_t = \left\{j : \mask_{t,j} = 1\right\}, \\
    \dataset&=\left[(\mathcal{X}_1, \mathcal{I}_1, \vec{y}_1), \dots, (\mathcal{X}_T,  \mathcal{I}_T, \vec{y}_T)\right].
\end{align}
The size of the input feature set $\mathcal{X}_t$ is time varying. To allow the model to determine which inputs are available, it is necessary to pass a set of feature IDs in an index set $\mathcal{I}_t$. Our proposed set learning module follows closely the ProtoRes architecture~\citep{oreshkin2022protores}. It takes the index set $\mathcal{I}_t$ and maps each of its active index positions to a continuous representation to create feature ID embeddings. It then concatenates each ID embedding to the corresponding feature value of $\mathcal {X}_t$. These feature values and ID embedding pairs are aggregated and summed to produce fixed dimensional vector representations that
we denote as $\vec x_0$.
The main components of our proposed set learning module are 
blocks, each consisting of $L$ fully connected (FC) layers. Residual skip connections are included in the architecture so that blocks can be bypassed.
An input set $\mathcal{X}_t$ is mapped to $\vec x_{0,t} = \textsc{EMB}(\mathcal{X}_t, \mathcal{I}_t)$. The overall structure of the module at block $r\in\{1, \dots,R\}$ (see Fig.~\ref{fig:block}) is (we are dropping time index for conciseness):
\begin{align}
    \vec h_{r,1} &= \textsc{FC}_{r,1}(\vec x_{r-1}),\, \dots,\, \vec h_{r,L} = \textsc{FC}_{r,L}(\vec h_{r,L-1}), \nonumber \\ 
    \vec x_r &= \textsc{ReLU}(\vec W_r\vec x_{r-1} + \vec h_{r-1,L}), \nonumber \\
    \hat{\vec y}_{r} &= \hat{\vec y}_{r-1} + \vec Q_L \vec h_{r,L}, \nonumber
\end{align}
where $\vec W_r, \vec Q_L$ are learnable matrices. We connect $R$ blocks sequentially to obtain the global output $\hat{\vec y}_R$.

\begin{table*}[t]
  \centering
   \caption{Comparison on standard datasets:
   cumulative error (mean $\pm$ st. deviation) over 20 runs. Bold indicates statistically significant result on Wilcoxon test $(p<0.05)$.
   }
  \begin{tabular}{l|ccccc}
  \hline
  \textbf{Dataset} & \textbf{OLVF} & \textbf{Aux-Drop (ODL)} & \textbf{Aux-Drop (OGD)} & \textbf{MODL} (ours)\\
\hline
  german & 333.4$\pm$9.7 & 306.6 $\pm$ 9.1 & 327.0$\pm$45.8 & \textbf{286.1 $\pm$ 5.3} \\ 
  svmguide3 & 346.4$\pm$11.6 & 296.9$\pm$1.3 & 296.6$\pm$0.6 & \textbf{288.0 $\pm$ 1.0} \\ 
  magic04 & 6152.4$\pm$54.7 &  5607.1  $\pm$ 235.1 &  5477.4 $\pm$ 299.3 & \textbf{5124.7 $\pm$ 153.4} \\
  a8a & 8993.8$\pm$40.3 & 6700.4$\pm$124.5 & 7261.8$\pm$283.5 & \textbf{5670.5 $\pm$ 278.2} \\ 
  \hline
  \end{tabular}
  \label{tab:uci}
\end{table*}
\begin{table*}[tb]
    \centering
    \caption{Comparison on HIGGS and SUSY for various feature probabilities $p_{\textrm{f}}$. Here, $p_{\textrm{f}}$ represents the i.i.d probability of a given feature being available during a time step. The metric is the mean ($\pm$ st. deviation) cumulative error in thousands over 5 runs. Bold indicates statistically significant Wilcoxon test $(p<0.05)$.}
    \label{tab:higgs+susy-key-results}
    \begin{tabular}[b]{c|cc}
    \hline
    \multirow{2}{*}{\textbf{$p_{\textrm{f}}$}} &  \multicolumn{2}{c}{\textbf{HIGGS}} \\ \cline{2-3}
    & Aux-Drop (ODL) & MODL (ours) \\
    \hline
    .01  &440.2 $\pm$ 0.1 &\textbf{439.6 $\pm$ 0.1}   \\
    .20  &438.4 $\pm$ 0.1 &\textbf{435.6 $\pm$ 0.4}   \\
    .50  &427.4 $\pm$ 0.7 &\textbf{422.7  $\pm$ 0.3}  \\
    .80  & 411.8 $\pm$ 0.4 &\textbf{399.6 $\pm$ 0.2}  \\
    .95  & 399.4 $\pm$ 1.0 &\textbf{377.1 $\pm$ 0.5} \\
    .99  & 392.0 $\pm$ 1.0 &\textbf{366.4 $\pm$ 0.5} \\
    \hline
    \end{tabular}
    \quad\quad
        \begin{tabular}[b]{c|cc}
    \hline
    \multirow{2}{*}{\textbf{$p_{\textrm{f}}$}} &  \multicolumn{2}{c}{\textbf{SUSY}} \\ \cline{2-3}
    & Aux-Drop (ODL) & MODL (ours) \\
    \hline
    .01  &285.0 $\pm$ 0.1 &\textbf{283.0 $\pm$ 0.1}   \\
    .20  &274.8 $\pm$ 0.9 &\textbf{271.8 $\pm$ 0.1}   \\
    .50  &256.6 $\pm$ 1.0 &\textbf{252.0  $\pm$ 0.1}  \\
    .80  & 237.0 $\pm$ 0.7 &\textbf{230.6 $\pm$ 0.1}  \\
    .95  & 226.2 $\pm$ 0.4 &\textbf{217.5 $\pm$ 0.1} \\
    .99  & 222.3 $\pm$ 0.2 &\textbf{212.2 $\pm$ 0.2} \\
    \hline
    \end{tabular}
\end{table*}
\begin{table}[ht]
  \centering
   \caption{Multiclass experiments missclassification rate. MODL outperforms both on small (CIFAR-10, 50k samples) and large datasets (I-MNIST, 1M samples).}
  \begin{tabular}{l|cc}
  \hline
  \textbf{Experiment} & \textbf{Aux-Drop (ODL)} & \textbf{MODL}\\
  \hline
  CIFAR-10 &88.9$\pm$0.3\%&\textbf{66.1$\pm$0.1}\%\\
  I-MNIST 1M&8.2$\pm$0.2\%&\textbf{3.8$\pm$0.1}\%\\
  \hline
  \end{tabular}
  \label{tab:multiclass}
\end{table}
\begin{table*}[t]
\centering
\caption{Ablation study on merging constituent learner outputs. We compare our proposed score sum approach to (i) mixture of experts (MoE); (ii) multiplying the learner logit probabilities; (iii) ensemble learning; (iv) an ad-hoc greedy weighting scheme that assigns more weight to learners with higher sliding window accuracy.
HIGGS ($p_{\textrm{f}}$=0.5), SUSY ($p_{\textrm{f}}$=0.99); $p_{\textrm{f}}$ represents the i.i.d probability of a given feature being available during a time step. 
}\label{tab:merge}
\begin{tabular}{l|ccccc}
\hline
\textbf{Dataset} & \textbf{Greedy Weighing} & \textbf{Multiplication} & \textbf{Ensemble} & \textbf{Mix. of Experts} & \textbf{Score Sum} (ours) \\ \hline
german & 293.0$\pm$10.3 & 294.9$\pm$7.1 & 307.5$\pm$21.0 & 316.2$\pm$9.3 & \textbf{285.9$\pm$7.2} \\
svmguide3 & 296.3$\pm$1.5 & 299.5$\pm$6.5 & 296.6$\pm$1.5 & 298.4$\pm$3.3 & \textbf{287.5$\pm$4.0} \\
magic04 & \textbf{4720$\pm$154} & 5146$\pm$75 & 6506$\pm$ 588 & 6719$\pm$44 & 5226 $\pm$ 98 \\
a8a & 6495$\pm$709 & 5691$\pm$40 & 5865$\pm$40 & 6190$\pm$168 & \textbf{5673$\pm$36} \\
HIGGS & 442.8$\pm$0.3k & 422.7$\pm$0.6k & 428.8$\pm$0.1k & 431.9$\pm$0.9k & \textbf{422.7$\pm$0.3k} \\
SUSY & 220.2$\pm$0.8k & 212.3$\pm$0.1k & 218.7$\pm$0.2k & 217.0$\pm$0.2k & \textbf{212.2$\pm$0.2k} \\ \hline
\end{tabular}
\end{table*}

\section{EXPERIMENTS}
\begin{figure*}[t]
    \centering
    \hspace{-2em}
    \vspace{-2cm}
    \rotatebox[origin=c]{90}{\bfseries Missclassification  Rate \strut}
        \includegraphics[width=0.29\textwidth, height=0.23\textwidth]{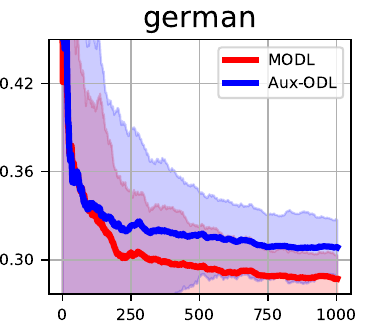}\,
        \includegraphics[width=0.29\textwidth, height=0.23\textwidth]{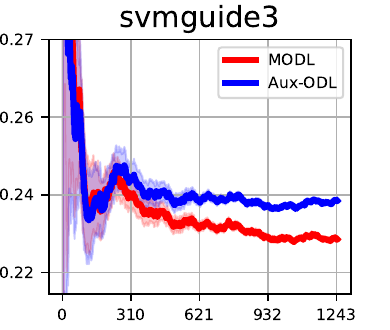}\,
        \includegraphics[width=0.29\textwidth, height=0.23\textwidth]{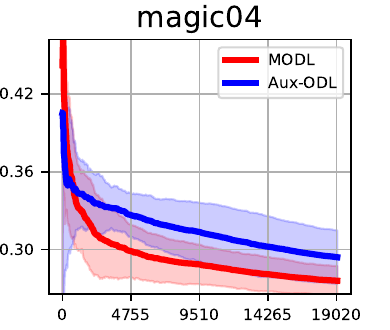}
        \includegraphics[width=0.29\textwidth, height=0.23\textwidth]{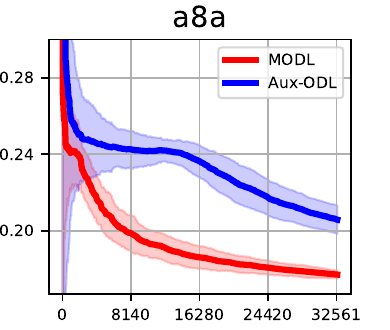}\,
        \includegraphics[width=0.29\textwidth, height=0.23\textwidth]{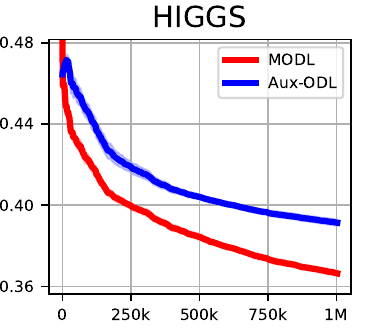}\,
        \includegraphics[width=0.29\textwidth, height=0.23\textwidth]{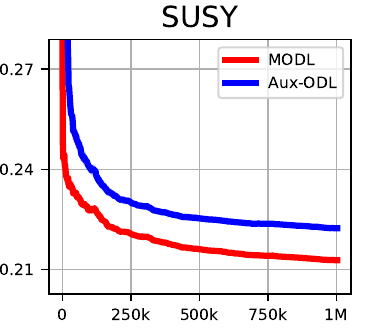}
        \rotatebox[origin=c]{0}{\bfseries Training Step \strut}
    \caption{Comparison of missclassification rate (lower is better) as a function of online training step for Aux-Drop (ODL)~\citep{agarwal2023auxdrop} (blue) vs. our proposed model MODL (red). Shaded regions indicate 95\% C.I.}
    \label{fig:learning_curves}
\end{figure*}
Our experiments provide empirical support to the following: (i) our online learning model MODL consistently outperforms state-of-the-art online deep learning methods; (ii) new proposed modules, including the online logistic regression as well as the set learning component, work synergistically within the MODL framework; (iii) combining the learners is best done via summation of unnormalized class scores, as opposed to alternatives, such as a mixture of experts or regular ensembles; (iv) our optimization framework is significantly more time efficient than hedge backpropagation based Aux-Drop (ODL) by~\citet{agarwal2023auxdrop}, which is the current best model.

\textbf{Datasets}. 
We employ established online deep learning benchmarks, replicating the setup from prior work by \cite{agarwal2023auxdrop}. Our analysis covers eight datasets of varying sizes. As noted in the review section, particle physics colliders generate vast amounts of data, necessitating the ability to learn from streaming data. Consequently, we selected two large particle physics datasets: \emph{HIGGS} and \emph{SUSY}~\citep{baldi2014data}. The \emph{HIGGS} dataset focuses on detecting the HIGGS boson in particle accelerator experiments, while \emph{SUSY} aims to identify SUper SYmmetric (SUSY) particles, distinguishing signal from background noise, which is particularly challenging.
In addition, we evaluate our model on the \emph{german} dataset~\citep{chang2011data}, which assesses consumer credit risk. We also include \emph{svmguide3}~\citep{dua2017uci}, a synthetic binary classification dataset. Another physics dataset, \emph{magic04}~\citep{chang2011data}, focuses on detecting gamma particle radiation in Cherenkov telescope images.  \emph{a8a}~\citep{dua2017uci} uses census data to predict individuals with high incomes based on demographic factors.
We also test on standard image classification datasets including
\emph{Infinite-MNIST} (I-MNIST)~\citep{loosli2007imnist} and \emph{CIFAR-10}~\citep{krizhevsky2012imagenet}.
Details about the datasets can be found in~\Cref{app:data}.

\textbf{Baselines.} We compare to standard methods from the literature. For experiments with unreliable features the current standard methods are OLVF~\citep{beyazit2019OLVF} and two versions of Aux-Drop~\citep{agarwal2023auxdrop}. The first version uses a fixed architecture and online gradient descent (ODG) updating~\citep{zinkevich2003OGD}, and the second uses ODL~\citep{sahoo2018ODL} hedge updates to learn the architecture jointly with the parameters. The latter variant of Aux-Drop is considered state-of-the-art. For the larger datasets we compare directly with the best method as other baselines are not designed for large datasets and very deep models. Continual learning approaches such as work by~\citet{kirkpatrick2017} or~\citet{prabhu2020} are not applicable as we are training from scratch.

\textbf{Training Methodology}. We closely follow the experimental setup of~\citet{agarwal2023auxdrop}. For small and medium datasets we run 20 independent trials, whereas for the large datasets we run 5. The random input masks for missing features and network initializations are from the same seed to ensure fairness. For all methods at each training step we process a single instance, so the batch size is fixed to 1. 
Hyperparameter sensitivity analysis is studied in~\Cref{app:learning_rate,app:layers}. For baselines we use the best hyperparameters reported in the literature.
For missing feature experiments we randomly mask all features with some probability (except for the first two features that are always available). Note that this is done to replicate the setting of prior work for the purpose of comparing the algorithms; our algorithm does not require any features to be always available. 

\textbf{MODL Architecture}: We used the exact same architecture in all  experiments. Specifically, the online logistic regression learner is implemented identically for all datasets and the MLP has 3 hidden layers with 250 neurons. The set learner consists of 6 blocks with 3 layers per block. The width of each layer is set to 128 neurons. For example, CIFAR-10 the capacity of the online logistic, MLP and Set learners is 3072, 893750, 1447312 learnable parameters respectively.
For a detailed description of hyperparameters, see~\Cref{app:hparams}.

\begin{table*}[]
\caption{Ablation of proposed model components. We show that for each learner that we add performance improves. The best model is our proposed MODL which uses all 3 components: online logistic regression (OLR), MLP and a set learner; $p_{\textrm{f}}$ represents the i.i.d probability of a given feature being available during a time step.}
\centering
\begin{tabular}{c|ccc|ccc}
\hline
\multirow{2}{*}{\textbf{$p_{\textrm{f}}$}} & \multicolumn{3}{c|}{\textbf{HIGGS}} & \multicolumn{3}{c}{\textbf{SUSY}} \\ \cline{2-7} 
 & OLR + MLP & OLR + Set & MODL (ours) & OLR + MLP & OLR + Set & MODL (ours) \\ \hline
.01 & 442.7$\pm$0.2 & 450.2$\pm$0.4 & \textbf{439.6 $\pm$ 0.1} & 285.3$\pm$0.1 & 332.6$\pm$0.3 & \textbf{283.0 $\pm$ 0.1} \\
.20 & 439.9$\pm$0.1 & 450.3$\pm$0.2 & \textbf{435.6 $\pm$ 0.4} & 274.0$\pm$0.1 & 352.4$\pm$0.3 & \textbf{271.8 $\pm$ 0.1} \\
.50 & 430.3$\pm$0.2 & 450.1$\pm$0.1 & \textbf{422.8 $\pm$ 0.3} & 255.0$\pm$0.1 & 343.3$\pm$0.6 & \textbf{252.0 $\pm$ 0.1} \\
.80 & 411.7$\pm$0.1 & 449.6$\pm$0.2 & \textbf{399.6 $\pm$ 0.2} & 234.3$\pm$0.1 & 326.3$\pm$0.4 & \textbf{230.6 $\pm$ 0.1} \\
.95 & 394.1$\pm$0.5 & 448.5$\pm$0.1 & \textbf{377.1 $\pm$ 0.5} & 222.1$\pm$0.1 & 320.9$\pm$0.4 & \textbf{217.5 $\pm$ 0.1} \\
.99 & 386.6$\pm$0.1 & 447.1$\pm$0.1 & \textbf{366.4 $\pm$ 0.6} & 217.3$\pm$0.1 & 316.3$\pm$0.6 & \textbf{212.2 $\pm$ 0.2} \\ \hline
\end{tabular}\label{tab:ablation}
\end{table*}
\textbf{Key Results}. \Cref{fig:learning_curves} shows that during training MODL remains consistently ahead of the current state-of-the-art model Aux-Drop (ODL) with respect to the classification error rate. Besides converging faster, our approach also achieves a lower overall error rate at the end of training. These findings are replicated across datasets that vary in size over 3 orders of magnitude and with different feature missingness levels. The detailed results for the cumulative miss-classification metric are summarized in~\Cref{tab:uci,tab:higgs+susy-key-results}. For the small and medium datasets shown in~\Cref{tab:uci} we can see that MODL is the clear top performer, reducing error on average by 11\%. We note that the improvements are measured over 20 trials and are all statistically significant at the 0.05 level using a paired Wilcoxon test. For the large datasets in~\Cref{tab:big_exp_full} we conduct a detailed feature missingness experiment for 13 different noise levels with 5 trials per level. The results show that MODL is consistently improving the state-of-the-art for all noise levels. This is an important result as it verifies the strength of our approach for a wide spectrum of feature noise, from near-zero up to extreme levels. We also validate our approach on multiclass image problems including the I-MNIST dataset with one million images and CIFAR-10 in~\Cref{tab:multiclass} where we achieve top performance. Note that on the complex but small CIFAR-10 dataset, Aux-Drop (ODL) struggles to learn anything in 50,000 steps (size of the dataset) as it has to both learn the architecture and the parameters of the early exit classifiers. On the other hand MODL quickly converges to reasonable performance. This is also reflected in large scale experiments in~\Cref{fig:learning_curves} such as HIGGS. Here, MODL has reasonable performance at 50,000 steps whereas Aux-Drop (ODL) has barely improved over untrained performance.

The learning efficiency of MODL optimization compared to Aux-Drop (ODL) extends past faster convergence. Aux-Drop (ODL), and base ODL operate at a higher training time complexity. The time complexity of performing a backpropagation update for a network with $L$ layers is $O(L^2)$ for standard implementations of ODL and Aux-Drop (ODL). This is analyzed in~\Cref{ssec:fast-odl},
where it is also empirically verified. On the other hand, backpropagation updates in MODL cost $O(L)$.
In our large scale experiments, compared to the AuxDrop (ODL) algorithm, our proposed MODL optimization reduces the the training time dramatically, \emph{e.g.}, from 63 hours to 9 when training with MLPs with the same number of layers and embedding dimension.

\textbf{Validation of MODL latent score summing.} The second ablation experiment explores different ways to merge learner predictions. MODL merges predictions by direct summation of the constituent learners' latent class scores. This is an important ablation result as it validates the MODL architecture and shows that letting the learners weigh their prediction individually works better than standard approaches. 
We consider alternative approaches that include: (i) learnable gating functions in a Mixture of Experts (MoE) style setup where the final output is a weighted sum of each learner's prediction; (ii) multiplying the logit probabilities (by summing the logarithm of the predicted logits); (iii) regular ensemble learning; and (iv) a greedy weighting scheme that assigns more weight to learners with higher sliding window accuracy. Our results in~\Cref{tab:merge} demonstrate the strength of latent score aggregation as an effective co-learning approach in MODL. Baseline implementation details are available in \Cref{app:details}.

\textbf{Ablation Studies}. In this section we rigorously verify our model components. 
\Cref{tab:ablation} shows that removing any individual component negatively affects the overall performance. From these results we ascertain that the presence of diverse learners with respect to the bias-variance trade-off bolsters performance. Additionally, in \Cref{tab:ablation_glr} we provide the results of an ablation study that investigates the use of online closed form updates versus a standard gradient based updating scheme for the logistic regression learner. Here, G-LR (gradient-based logistic regression) is the control model where the closed form solution is replaced by backpropogation. We keep all parameters the same as in the original experiment. We set the learning rate for the gradient-based logistic regression updates to 0.01 (small datasets) and 0.001 (larger datasets). These values are the same leaning rates as used for the other gradient-based algorithms in MODL.
Comparing the closed form solution MODL (ours) with the gradient-based solution MODL (G-LR)  demonstrates the benefit of the closed form solution.
\begin{table}[]
\centering
\caption{Ablation of proposed online update scheme for the logistic regression learner. G-LR (gradient-based logistic regression) is the control model where the closed form solution is replaced by backpropogation. We conduct 100 and 25 runs for the small and large datasets respectively.}
\begin{tabular}{@{}lcc@{}}
\toprule
\textbf{Dataset} & \textbf{MODL (G-LR)} & \textbf{MODL (ours)} \\ \midrule
german 	&325.2 $\pm$ 8.9 	&\textbf{289.4 $\pm$5.4}\\
svmguide3 	& 293.0$\pm$ 3.3 	&\textbf{287.5 $\pm$3.6}\\
magic04 	&5652.5 $\pm$51.9 	&\textbf{5137.1 $\pm$53.5}\\
a8a 	&5829.6 $\pm$27.1 	&\textbf{5493.2 $\pm$51.2}\\
HIGGS 	&384.4 $\pm$0.2 	&\textbf{362.3 $\pm$0.2}\\
SUSY 	&214.0 $\pm$0.1 	&\textbf{210.5 $\pm$0.2}\\ \bottomrule
\end{tabular}\label{tab:ablation_glr}
\end{table}

\textbf{Robustness to hyperparameter selection.} A particularly challenging and sensitive hyperparameter to select in online learning is the learning rate. We run sensitivity experiments with respect to the learning rate in the large benchmarks. Our results in~\Cref{app:learning_rate} show increased robustness to learning rate selection. This is expected because our framework incorporates learners that train without backpropagation, which safeguards against selecting learning rates that are far from optimal. Our framework MODL may work with as many learners as the user wishes to implement. In the main experiments we proposed a standard setup that achieves practically good results with 3 learners. However, this does not preclude the user from introducing more than 3 learners. In the next experiment we implement a 5 learner setup by introducing MLPs with additional layers to obtain learners with different characteristics along the bias-variance tradeoff. We investigate the robustness of MODL to the number of learners in~\Cref{tab:five_learners}.
\begin{table}[]
\centering
\caption{MODL demonstrates low sensitivity to number of learners selected, showing robustness to hyperparameter selection. We observe that MODL performance is stable when we use three or five learners. The two additional learners are higher capacity MLPs.}
\begin{tabular}{@{}lcc@{}}
\toprule
\textbf{Dataset} & \textbf{3 models MODL} & \textbf{5 models MODL} \\ \midrule
german & 286.1 $\pm$ 5.3 & 290.1 $\pm$ 7.3 \\
svmguide3 & 288.0 $\pm$ 1.0 & 287.0 $\pm$ 4.1 \\
magic04 & 5124.7 $\pm$ 153.4 & 5239.7 $\pm$ 175.2 \\
a8a & 5670.5 $\pm$ 278.2 & 5639.3 $\pm$ 144.2 \\ \bottomrule
\end{tabular}\label{tab:five_learners}
\end{table}

\section{CONCLUSION}
\textbf{Limitations}. 
Our work is focused on supervised learning from streams of data. This is not trivially applicable to other areas such as online reinforcement learning. While our work shows that including multiple learners with different bias-variance trade-offs is a sensible approach to online deep learning, it does not come with theoretical guarantees for error or regret bounds.

\textbf{Overview}. We demonstrate that the problem of online deep learning is best handled synergistically by multiple learners that operate at various points of the convergence speed/expressivity trade-off curve. We derive a very fast learning and non-backpropagation based approach to support early stages of learning and to provide a strong baseline for other learners. Then, at the other end of the spectrum, we propose a slowly learning, yet very expressive set learner that is invariant to feature ordering and robust to missing features. We show that the appropriate way to combine these learners is a novel stacking regime where the models learn to predict on top of one-another rather than ensembling or via a mixture of experts. Our approach significantly improves performance and reduces training complexity.

\subsubsection*{Acknowledgements}
This research was funded by the Natural Sciences and Engineering Research Council of Canada (NSERC), [reference number 260250].
Cette recherche a été financée par le Conseil de recherches en sciences naturelles et en génie du Canada (CRSNG), [numéro de référence 260250]. 
During this research project Antonios Valkanas was supported through NSERC Postgraduate Scholarships – Doctoral (PGS-D) program, the Stavros S. Niarchos Foundation Fellowship and the Vadasz Doctoral Fellowship.

\setcitestyle{numbers} %
\bibliographystyle{plainnat}
\bibliography{refs}
\clearpage

\section*{Checklist}

 \begin{enumerate}

 \item For all models and algorithms presented, check if you include:
 \begin{enumerate}
   \item A clear description of the mathematical setting, assumptions, algorithm, and/or model. [Yes]
   \item An analysis of the properties and complexity (time, space, sample size) of any algorithm. [Yes]
   \item (Optional) Anonymized source code, with specification of all dependencies, including external libraries. [Not Applicable]
   
   Code will be made public upon paper acceptance.
 \end{enumerate}

 \item For any theoretical claim, check if you include:
 \begin{enumerate}
   \item Statements of the full set of assumptions of all theoretical results. [Yes]
   \item Complete proofs of all theoretical results. [Yes]
   \item Clear explanations of any assumptions. [Yes]     
 \end{enumerate}

 \item For all figures and tables that present empirical results, check if you include:
 \begin{enumerate}
   \item The code, data, and instructions needed to reproduce the main experimental results (either in the supplemental material or as a URL). [Yes/No/Not Applicable]
   \item All the training details (e.g., data splits, hyperparameters, how they were chosen). [Yes]
         \item A clear definition of the specific measure or statistics and error bars (e.g., with respect to the random seed after running experiments multiple times). [Yes]
         \item A description of the computing infrastructure used. (e.g., type of GPUs, internal cluster, or cloud provider). [Yes]
 \end{enumerate}

 \item If you are using existing assets (e.g., code, data, models) or curating/releasing new assets, check if you include:
 \begin{enumerate}
   \item Citations of the creator If your work uses existing assets. [Yes]
   \item The license information of the assets, if applicable. [Yes]
   \item New assets either in the supplemental material or as a URL, if applicable. [Yes]
   \item Information about consent from data providers/curators. [Not Applicable]
   \item Discussion of sensible content if applicable, e.g., personally identifiable information or offensive content. [Not Applicable]
 \end{enumerate}

 \item If you used crowdsourcing or conducted research with human subjects, check if you include:
 \begin{enumerate}
   \item The full text of instructions given to participants and screenshots. [Not Applicable]
   \item Descriptions of potential participant risks, with links to Institutional Review Board (IRB) approvals if applicable. [Not Applicable]
   \item The estimated hourly wage paid to participants and the total amount spent on participant compensation. [Not Applicable]
 \end{enumerate}

 \end{enumerate}

 \newpage
 \onecolumn
\appendix

\addcontentsline{toc}{section}{Appendix} %
\part{\centerline{MODL: Multilearner Online Deep Learning}\centerline{Supplementary material}} %
\parttoc %
\newpage
\section{Broader Impact Statement}\label{app:impact}

Our paper introduces a new Online Learning technique and improves existing training methodologies for online learning of deep neural networks. We show that co-training multiple learners can lead to significantly faster convergence as well as improved overall model performance at inference time. 

A strongly positive outcome stemming from our contributions is the significant reduction of training time from quadratic complexity in network parameters down to linear complexity. This has profound consequences for training deep models online as it significantly decreases the necessary training compute. Thus the energy spent for training deep online learners is massively reduced. 

Furthermore, the improved convergence speed and overall performance of our proposed technique MODL means that our model is more data efficient than existing models, leading to a moderately reduced need for large dataset sizes. Efficiently learning in one pass means that data does not need to be stored, which can protect the privacy rights of individuals and organizations. Training online without storing data can help companies comply with data protection laws and allow the consumer to have greater confidence that their fundamental human right of privacy is protected.

We are confident that our research effort offers more benefits in energy saving and data privacy as opposed to the risks posed by the usage of deep models. Furthermore, we point out that the risks posed by a potential deployment of our model can be hedged against due to the interpretability of some of our constituent learners. In our case, the potential to interpret model outputs is particularly strong for the weaker learners such as logistic regression.
\newpage

\section{Additional Background}
\subsection{Model weighing approaches}
Suppose we have a set of trained models. 
It is frequently suboptimal to discard all models except the best performing one. Approaches such as Bayesian model averaging (BMA) estimate the predicted quantity under each candidate model and then produce a weighted average estimate over all models. In BMA, the weights are determined according to the probability that each model represents the true data generating mechanism (DGM) given the data~\citep{wilson2020bayesian}. However, BMA assumes that at least one element of the model set we are averaging over contains a correctly specified model that can truly encompass the DGM~\citep{wasserman2000bma}. This assumption is not true in general deep learning settings; this often leads to suboptimal generalization of BMA~\citep{masegosa2020misspecification}. We therefore search for alternative methods for building fundamental model architectures capable of combining multiple learners. One such approach is model stacking~\citep{wolpert1992stacking, gneiting2007strictly}. While stacking was originally only capable of producing point estimates, renewed interest has recently framed it from a Bayesian perspective~\citep{clydec2013bayesian, le2017stacking}. Recent approaches cast selection of stacking weights as a Bayesian decision problem~\citep{yao2024stacking}.

\subsection{Online Learning with Missing Features}
In this section we describe in detail the standard base architecture that underpins most modern online deep learning techniques.
\Cref{alg:odl} outlines the current online deep learning state-of-the-art base architecture, ODL, developed by~\citet{sahoo2018ODL} and refined in Aux-Drop (ODL)~\citep{agarwal2023auxdrop}. This section reviews this architecture, which is based on a joint bi-level optimization objective. The learning (or selection) of the architecture is accomplished by attaching early exit classifiers to each hidden layer, and taking a weighted convex sum of all classifiers to produce the overall output. The weights attached to each classifier's output in the sum are defined as $\alpha$ and the network architecture is selected by optimizing these $\alpha$ values. Each alpha connects one of the hidden layers with the overall neural network output. The other optimization level handles the optimization of model parameters $\Theta$. The algorithm oscillates between taking an optimization step in the network weights and a ``hedge''~\citep{freund1997backprop} optimization step that effectively updates the architecture by assigning weights to the skip connections  (see last step of~\Cref{alg:odl}).

Consider a deep neural network with $L$ hidden layers where each layer is connected to an early exit predictor $\vec f^{(i)}$. The prediction function $\vec F$ for the deep neural network is given by:
\begin{align}\label{eq:hdnn}
\vec{F}(\vec{x}) &= \sum_{l=0}^L \alpha^{(l)} \vec{f}^{(l)}\,,\quad \text{where} \,\,\,  \vec{f}^{(l)}   = \mathrm{softmax}(\vec{h}^{(l)} \boldsymbol{\Theta^{(l)})},\\
\nonumber   \vec{h}^{(l)}  &= \sigma(\mathbf{W}^{(l)} \vec{h}^{(l-1)}),\ \forall l = 1,\dots, L; 
\nonumber  \vec{h}^{(0)}  = \vec{x} \in \mathbb{R}^{d_\text{in}}.
\end{align}
Here, $\boldsymbol{\Theta^{(l)}}$ is the parameter matrix of the early exit classifier $\vec{f}^{(l)}$, and  $\mathbf{W}^{(l-1)}$ is the parameter matrix of the hidden layer that yields intermediate representation $\vec{h}^{(l)}$.
Learning the parameters $\boldsymbol{\Theta^{(l)}}$ for  classifiers $\vec{f}^{(l)}$ can be done via online gradient descent (OGD), where the input to the $l^{th}$ classifier is $\vec{h}^{(l)}$. This is essentially standard backpropagation  with learning rate $\eta$:
\begin{align}\label{eq:classifierUpdate}
\boldsymbol{\Theta}^{(l)}_{t+1}  &\leftarrow  \boldsymbol{\Theta}^{(l)}_t - \eta \nabla_{\boldsymbol{\Theta}^{(l)}_t} \loss(\vec{F}(\vec{x}_t, y_t)). %
\end{align}
Updating the feature representation parameters $\mathbf{W}^{(l)}$ potentially requires backpropagating through all classifiers $\f{(l)}$. Thus, using the adaptive loss function, $\loss(\mathbf{F}(\vec{x}), y) = \sum_{l=0}^L \alpha^{(l)}\loss(\f{(l)}(\vec{x}), y)$, and applying OGD, the update rule for $\mathbf{W}^{(l)}$ is given by:
\begin{align}\label{eq:hdnnBackprop}
\mathbf{W}^{(l)}_{t+1}  \leftarrow \mathbf{W}^{(l)}_{t} - \eta \sum_{j=l}^L \alpha^{(j)} \nabla _{\mathbf{W}^{(l)}} \loss(\f{(j)}, y_t)\,,
\end{align}
where $\nabla _{\mathbf{W}^{(l)}} \loss(\f{(j)}, y_t)$ is computed via backpropagation from error derivatives of $\vec{f}^{(j)}$. The summation in the gradient term starts at $j=l$ because the shallower classifiers do not depend on $\mathbf{W}^{(l)}$ for making predictions. This training technique is summarized in Algorithm~\ref{alg:odl}.

When~\cref{eq:hdnnBackprop} is implemented directly, a single backpropagation pass for this architecture incurs an $O(L^2)$ cost, which is evident from its implementation code\footnote{The most popular ODL implementation has 171 stars and 44 forks at the time of writing and has quadratic training complexity: \url{https://github.com/alison-carrera/onn}}. We also demonstrate this in the proof of Proposition~\ref{prop:backprop}. This can grow prohibitively expensive for deep networks and has had major implications in the efficiency of subsequent research methods and applications that rely on ODL. For example, works as recent as Aux-Drop~\citep{agarwal2023auxdrop} build on this framework.
In the next section, we first show how an equivalent update to~\cref{eq:hdnnBackprop} can be derived by ``rewiring'' the network to reduce time complexity from $O(L^2)$ to $O(L)$ per backpropagation update. \textbf{This improved implementation of the base ODL architecture constitutes a secondary yet notable contribution of our work that we did not explore in the main text due to space limitations.} We carefully derive and prove the improved complexity of our implementation in the subsequent section.
\section{Fast Online Deep Learning} \label{ssec:fast-odl}
\textbf{Fast ODL (FODL)}. During training, rather than backpropagating through $L$ individual early exit classifier losses, we propose aggregating the outputs and backpropagating through a linear combination to obtain $\mathcal{L}_{\textsc{tot}}$. Detailed architecture  schematics and analysis of this are provided in~\Cref{app:backprop}. This change in the network topology has important ramifications for training via backpropagation.
In particular, this alters~\cref{eq:hdnnBackprop} as follows:
\begin{align}
    W^{(l)}_{t+1}  &\leftarrow W^{(l)}_{t} - \eta \nabla_{W^{(l)}} \loss_{\textsc{tot}}, \quad \text{where} \label{eq:new_grad}\\
    \loss_{\textsc{tot}} &= \left[\alpha^{(1)}, \dots, \alpha^{(L)}\right]\left[\loss(\f{(1)}, y_t), \dots, \loss(\f{(L)}, y_t) \right]^T = \sum_{j=1}^L \alpha^{(j)} \loss(\f{(j)}, y_t)\,. \label{eq:tot_loss}
\end{align}

\begin{prop}\label{prop:backprop}
    The updates described by~\cref{eq:hdnnBackprop} and~\cref{eq:new_grad} produce the same gradient update
    but the proposed training scheme has lower complexity. For a dataset of size $n$ and a network with $L$ layers,~\cref{eq:hdnnBackprop} has training time complexity (due to backpropagation) $O(nL^2)$, but~\cref{eq:new_grad} only requires $O(nL)$ computation. 
\end{prop}
\begin{proof}
    The proof is in~\Cref{app:backprop}.
\end{proof}

\newpage
\section{Algorithms}\label{app:algorithms}

\begin{algorithm}[]
	\caption{Online Deep Learning (ODL) using Hedge Backpropagation}\label{alg:odl}
\begin{algorithmic}
  \Require 
	Learning rate Parameter: $\eta$, discount parameter $\beta$\\
	\textbf{Initialize}: {$F(\vec{x})$ with $L$ hidden layers and $L+1$ classifiers $f^{(l)}$; $\alpha^{(l)} = \frac{1}{L+1},  \forall l = 0, \dots, L$}\\
 \For{t = 1,\dots,T}
		\State Receive instance $\vec{x}_t$ and predict $\hat{y_t} = F_t(\vec{x}_t) = \sum_{l=0}^L \alpha_t^{(l)} f_t^{(l)} $ via \cref{eq:hdnn}
		\State Reveal true value $y_t$ and calculate $\loss^{(l)}_t = \loss(\f{(l)}_t(\vec{x}_t),y_t), \forall l, \dots, L$;
		\State Update $\Theta^{(l)}_{t+1}, \forall l = 0, \dots, L$  via \cref{eq:classifierUpdate} and $W^{(l)}_{t+1}, \forall l = 1, \dots, L$ via \cref{eq:hdnnBackprop};
		\State Update $\alpha^{(l)}_{t+1} = \alpha^{(l)}_{t} \beta^{\loss^{(l)}_t}, \forall l = 0, \dots, L$ and normalize $\alpha^{(l)}_{t+1}$ to sum to 1.
		\EndFor
	\end{algorithmic}
\end{algorithm}
\bigskip
\begin{algorithm}[]
	\caption{Online Bayesian Logistic Regression}
\begin{algorithmic}
  \Require Dataset $\dataset = \{(\vec x_i, y_i)\}_{i=1}^n$\\
  \textbf{Initialize}: $p(\theta)=\mathcal{N}(\theta; \vec m_0, \vec P_0)$, $\vec m_0 = 0$, where $\mathbf{P_0} = \mathbf{I}$\\
  \textbf{Return} $p(\theta | \{(\vec x_i, y_i)\}_{i=1}^n) = \mathcal{N}(\theta; \vec m_n, \vec P_n)$
 \For{t = 1,\dots,n}
		\State Receive instance: $\vec{x}_t$ and predict: $\hat{y}_t = \sigma(\vec x_t\vec m_{t-1})$
		\State Reveal true value: $y_t$ and update parameters: 
          \begin{align}
            \Xi_t &= \vec x_t \vec P_{t-1} \vec x_t^\top + \vec P \left[\left(1-\vec \sigma(\vec x_t \vec m_{t-1})\right) \sigma\left(\vec x_t \vec m_{t-1}\right)\right]^2,\\
            \vec K_t &=  \vec P_{t-1} \vec X_t^\top \left(1-\vec \sigma(\vec x_t \vec m_{t-1})\right) \sigma\left(\vec x_t \vec m_{t-1}\right)\Xi_t^{-1},\\
            \vec m_t &= \vec m_{t-1} + \vec K_t [\vec y_t - \sigma\left(\vec x_t \vec m_{t-1}\right)],\\
            \vec P_{t} &=  \vec P_{t-1} - \vec K_t \Xi_t \vec K_t^\top.
        \end{align}
		\EndFor
	\end{algorithmic}
\end{algorithm}

\newpage

\section{Additional Experiments}
In this section we provide additional experiments to verify four research claims empirically.
\begin{enumerate}
    \item MODL outperforms Aux-Drop (ODL) for all feature missingness settings on large benchmarks (HIGGS, SUSY).
    \item Removing any model component hurts model performance for all missingness levels on large benchmarks. This validates the choice of using multiple learners across the bias-variance trade-off spectrum.
    \item Our improved implementation of ODL and Aux-Drop ODL, which we call Fast ODL, and abbreviate as FODL and Aux-Drop FODL, has lower time complexity without sacrificing any performance compared to the standard ODL and Aux-Drop (ODL) implementation.
    \item Our proposed approach MODL performs better for a wider range of learning rates. This sensitivity experiment is key as learning rate selection is one of the most difficult hyperparameters to tune in Online Learning.
\end{enumerate}
\subsection{Feature Missingness Experiment}
In this section we provide additional experiments in support of the summary results presented in the main text.
More concretely, we provide a full feature missningess study with 13 unique $p_{\textbf{f}}$ values for the large benchmarks HIGGS and SUSY. As shown in Tab.~\ref{tab:big_exp_full}, our model convincingly outperforms for all settings. We observe that the advantage of our method increases as more features become available.
\begin{table}[h]
    \centering
    \caption{Comparison on HIGGS and SUSY for various feature probabilities $p_{\textrm{f}}$. The metric is the mean ($\pm$ standard deviation) cumulative error in thousands (5 runs).}
    \label{tab:big_exp_full}
    \begin{tabular}[]{c|cc}
    \hline
    \multirow{2}{*}{\textbf{$p_{\textrm{f}}$}} &  \multicolumn{2}{c}{\textbf{HIGGS}} \\ \cline{2-3}
    & Aux-Drop (ODL) & MODL (ours) \\
    \hline
    .01  &440.2 $\pm$ 0.1 &\textbf{439.6 $\pm$ 0.1}   \\
    .05  &440.0 $\pm$ 0.1 &\textbf{439.5 $\pm$ 0.2}   \\
    .10  &440.0 $\pm$ 0.2 &\textbf{438.5 $\pm$ 0.1}   \\
    .20  &438.4 $\pm$ 0.1 &\textbf{435.6 $\pm$ 0.4}   \\
    .30  &435.1 $\pm$ 0.2 &\textbf{432.3 $\pm$ 0.3}   \\
    .40  &432.0 $\pm$ 0.3 &\textbf{428.4 $\pm$ 0.3}   \\
    .50  &427.4 $\pm$ 0.7 &\textbf{422.8 $\pm$ 0.4}  \\
    .60  &423.2 $\pm$ 0.5 &\textbf{422.8 $\pm$ 0.2}  \\
    .70  &418.5 $\pm$ 0.7 &\textbf{409.5 $\pm$ 0.4}  \\
    .80  &411.8 $\pm$ 0.4 &\textbf{399.6 $\pm$ 0.3}  \\
    .90  &405.6 $\pm$ 0.7 &\textbf{387.0 $\pm$ 0.3}  \\
    .95  &399.4 $\pm$ 1.0 &\textbf{377.2 $\pm$ 0.5} \\
    .99  &392.1 $\pm$ 1.0 &\textbf{366.5 $\pm$ 0.6} \\
    \hline
    \end{tabular}
    \quad\quad
        \begin{tabular}[]{c|cc}
    \hline
    \multirow{2}{*}{\textbf{$p_{\textrm{f}}$}} &  \multicolumn{2}{c}{\textbf{SUSY}} \\ \cline{2-3}
    & Aux-Drop (ODL) & MODL (ours) \\
    \hline
    .01  &285.0 $\pm$ 0.1 &\textbf{283.0 $\pm$ 0.1}   \\
    .05  &283.3 $\pm$ 0.2 &\textbf{281.0 $\pm$ 0.1}   \\
    .10  &280.6 $\pm$ 0.5 &\textbf{278.0 $\pm$ 0.1}   \\
    .20  &274.9 $\pm$ 0.9 &\textbf{271.9 $\pm$ 0.1}   \\
    .30  &269.0 $\pm$ 0.7 &\textbf{265.5 $\pm$ 0.1}   \\
    .40  &262.8 $\pm$ 0.9 &\textbf{258.9 $\pm$ 0.1}   \\
    .50  &256.7 $\pm$ 1.0 &\textbf{252.0 $\pm$ 0.1}  \\
    .60  &250.0 $\pm$ 0.9 &\textbf{244.7 $\pm$ 0.2}  \\
    .70  &243.9 $\pm$ 0.7 &\textbf{238.0 $\pm$ 0.2}  \\
    .80  &237.0 $\pm$ 0.7 &\textbf{230.6 $\pm$ 0.1}  \\
    .90  &230.0 $\pm$ 0.7 &\textbf{222.3 $\pm$ 0.2}  \\
    .95  &226.2 $\pm$ 0.4 &\textbf{217.5 $\pm$ 0.1} \\
    .99  &222.2 $\pm$ 0.2 &\textbf{212.2 $\pm$ 0.2} \\
    \hline
    \end{tabular}
\end{table}
\subsection{Model Ablation Study}
In this section we empirically validate our proposed model MODL. As shown in~\Cref{tab:ablation_learners} the results in the main paper persist in the additional feature missningness settings,  further validating our conclusion that all of our model components are necessary.
\begin{table}[h]
\caption{Ablation of proposed model components. We show that for each learner that we add the overall model performance improves. OLR refers to online logistic regression, ``Set'' is short for Set Learner model.}
\centering
\begin{tabular}{c|ccc|ccc}
\hline
\multirow{2}{*}{\textbf{$p_{\textrm{f}}$}} & \multicolumn{3}{c|}{\textbf{HIGGS}} & \multicolumn{3}{c}{\textbf{SUSY}} \\ \cline{2-7} 
 & OLR + MLP & OLR + Set & MODL (ours) & OLR + MLP & OLR + Set & MODL (ours) \\ \hline
.01 & 442.8$\pm$0.2 & 450.2$\pm$0.5 & \textbf{439.6 $\pm$ 0.1} & 285.3$\pm$0.1 & 332.6$\pm$2.1 & \textbf{283.0 $\pm$ 0.1} \\
.05 & 442.7$\pm$0.1 & 449.9$\pm$0.4 & \textbf{439.5 $\pm$ 0.2} & 283.3$\pm$0.1 & 334.7$\pm$1.9 & \textbf{281.1 $\pm$ 0.1} \\
.10 & 441.9$\pm$0.1 & 450.1$\pm$0.3 & \textbf{438.5 $\pm$ 0.1} & 280.0$\pm$0.1 & 337.8$\pm$1.4 & \textbf{278.0 $\pm$ 0.1} \\
.20 & 439.9$\pm$0.1 & 450.3$\pm$0.2 & \textbf{435.6 $\pm$ 0.4} & 274.0$\pm$0.1 & 352.4$\pm$0.3 & \textbf{271.9 $\pm$ 0.1} \\
.30 & 437.4$\pm$0.1 & 450.1$\pm$0.3 & \textbf{432.3 $\pm$ 0.3} & 267.9$\pm$0.1 & 355.4$\pm$0.3 & \textbf{265.5 $\pm$ 0.1} \\
.40 & 434.9$\pm$332 & 450.3$\pm$404 & \textbf{428.4 $\pm$ 302} & 261.7$\pm$0.1 & 338.9$\pm$0.4 & \textbf{258.9 $\pm$ 0.1} \\
.50 & 430.3$\pm$193 & 450.1$\pm$93 & \textbf{422.8 $\pm$ 386} & 255.0$\pm$0.1 & 343.3$\pm$0.7 & \textbf{252.0 $\pm$ 0.1} \\
.60 & 425.4$\pm$178 & 449.9$\pm$137 & \textbf{422.8 $\pm$ 244} & 248.3$\pm$0.1 & 336.6$\pm$0.5 & \textbf{244.7 $\pm$ 0.2} \\
.70 & 420.5$\pm$484 & 451.0$\pm$122 & \textbf{409.5 $\pm$ 424} & 241.3$\pm$0.1 & 337.9$\pm$0.4 & \textbf{238.0 $\pm$ 0.1} \\
.80 & 411.8$\pm$106 & 449.6$\pm$219 & \textbf{399.6 $\pm$ 256} & 234.3$\pm$0.1 & 326.3$\pm$0.4 & \textbf{230.6 $\pm$ 0.1} \\
.90 & 401.4$\pm$517 & 449.8$\pm$58 & \textbf{387.0 $\pm$ 348} & 226.5$\pm$0.2 & 322.0$\pm$0.4 & \textbf{222.3 $\pm$ 0.2} \\
.95 & 394.1$\pm$565 & 448.5$\pm$175 & \textbf{377.2 $\pm$ 537} & 222.1$\pm$0.1 & 320.9$\pm$0.4 & \textbf{217.5 $\pm$ 0.1} \\
.99 & 386.7$\pm$133 & 447.1$\pm$163 & \textbf{366.4 $\pm$ 588} & 217.4$\pm$0.1 & 316.4$\pm$0.7 & \textbf{212.2 $\pm$ 0.2} \\ \hline
\end{tabular}\label{tab:ablation_learners}
\end{table}

\newpage

\subsection{Training Time Additional Settings}\label{app:layers}
In this section, we explore the wall clock training time reduction offered by our proposed fast training scheme on large datasets. We note that the training time savings range from 30\% to over 80\%.
\begin{table}[h]
  \centering
   \caption{Time comparison between Aux-DropODL and FastAux-DropODL on HIGGS and SUSY for $p=0.5$ (various layer and embedding dimensions).}
  \begin{tabular}{l|cccc}
  \hline
  \textbf{Experiment} & \textbf{Aux-Drop (ODL)} & \textbf{Aux-Drop (FODL)} &\textbf{Time ODL vs. FODL} \\
  \hline
  HIGGS ($L=5, E=25$)&431.9 $\pm$ 0.5&431.4 $\pm$ 0.5&5:35:10 vs. 3:34:55\\
  HIGGS ($L=5, E=50$)&429.6 $\pm$ 0.5&429.6 $\pm$ 0.6&5:41:05 vs. 3:38:18\\
  HIGGS ($L=5, E=100$)&428.0 $\pm$ 0.3&427.9 $\pm$ 0.3&5:45:56 vs. 3:38:42\\
  \hline
  HIGGS ($L=11, E=25$)&429.8 $\pm$ 0.7&429.7 $\pm$ 0.4&22:44:06 vs. 6:21:58\\
  HIGGS ($L=11, E=50$)&427.4 $\pm$ 0.5&427.4 $\pm$ 0.7&16:04:59 vs. 4:29:05\\
  HIGGS ($L=11, E=100$)&426.9 $\pm$ 0.3&426.6 $\pm$ 0.3&20:05:39 vs. 6:32:20\\
  \hline
  HIGGS ($L=20, E=25$)&431.4 $\pm$ 0.5&431.0 $\pm$ 0.5&57:04:12 vs. 9:23:13\\
  HIGGS ($L=20, E=50$)&429.2 $\pm$ 0.4&429.4 $\pm$ 0.4&54:04:10 vs. 8:43:11\\
  HIGGS ($L=20, E=100$)&427.6 $\pm$ 0.5&427.8 $\pm$ 0.4 &63:19:05 vs. 8:58:51\\
  \hline
  SUSY ($L=5, E=25$)&257.3 $\pm$ 1.0&257.3 $\pm$ 1.2& 5:33:53 vs. 2:59:32\\
  SUSY ($L=5, E=50$)&256.6 $\pm$ 0.9&256.6 $\pm$ 0.9& 5:37:42 vs. 3:03:09\\
  SUSY ($L=5, E=100$)&255.9 $\pm$ 0.4&255.8 $\pm$ 0.5&5:07:21 vs. 3:41:03\\
  \hline
  SUSY ($L=11, E=25$)&257.2 $\pm$ 0.9&257.5 $\pm$ 1.1&22:41:05 vs. 6:19:03\\
  SUSY ($L=11, E=50$)&257.0 $\pm$ 1.2&256.7 $\pm$ 1.0&23:04:11 vs. 6:54:21\\
  SUSY ($L=11, E=100$)&255.9 $\pm$ 0.8&256.1 $\pm$ 0.8&20:05:06 vs. 6:29:58\\
  \hline
  SUSY ($L=20, E=25$)&258.6 $\pm$ 1.1&258.5 $\pm$ 1.3& 36:39:13 vs. 8:44:24\\
  SUSY ($L=20, E=50$)&257.4 $\pm$ 0.9&257.4 $\pm$ 0.9& 39:59:38 vs. 8:01:53\\
  SUSY ($L=20, E=100$)&257.3 $\pm$ 1.0&257.1 $\pm$ 1.0& 43:07:45 vs. 8:19:44\\
  \hline
  \end{tabular}
  \label{tab:time_comp}
\end{table}

\subsection{Learning Rate Sensitivity Experiments}\label{app:learning_rate}
In this section we demonstrate that our model has stable and better performance than the baseline approaches for a broad range of learning rates. This is due to the incorporation of learners without a backpropagation update. This shields the overall architecture from poorer learning rate selections.
Note that if we set the learning rate too high the other learners may become unstable. Protection from bad learning rates thus has limits (i.e., cannot train with very fast learning rates such as 0.5). The conclusion from this section and Tab.~\ref{tab:lr_study} is that for any reasonable learning rate selection for deep models our model outperforms.

\begin{table}[]
\centering
\caption{Error in HIGGS and SUSY for various learning rates with fixed probability of unreliable features $p_f=0.99$. The metric is reported as the mean $\pm$ standard deviation of the number of errors in 5 runs (11 layer Aux networks).}
\begin{tabular}{@{}rcc|cc@{}}
\toprule
\multicolumn{1}{c}{\multirow{2}{*}{\textbf{lr}}} & \multicolumn{2}{c}{\textbf{HIGGS}} & \multicolumn{2}{c}{\textbf{SUSY}} \\ \cmidrule(l){2-5} 
\multicolumn{1}{c}{} & Aux-Drop (ODL) & MODL (ours) & Aux-Drop (ODL) & MODL (ours) \\ \midrule
\multicolumn{1}{r|}{0.00005} & 470.9$\pm$0.8 & \textbf{426.1$\pm$ 1.2} & 391.9 $\pm$ 67 & \textbf{225.5 $\pm$ 0.7} \\
\multicolumn{1}{r|}{0.0001} & 470.1$\pm$0.7 & \textbf{415.4$\pm$0.8} & 333.4$\pm$62 & \textbf{222.2 $\pm$ 0.4} \\
\multicolumn{1}{r|}{0.0005} & 442.8$\pm$3.5 & \textbf{392.7$\pm$0.6} & 278.6$\pm$21 & \textbf{216.6 $\pm$ 0.2} \\
\multicolumn{1}{r|}{0.001} & 402.1$\pm$6.4 & \textbf{366.5$\pm$0.6} & 225.9$\pm$1.9 & \textbf{212.2 $\pm$ 0.2} \\
\multicolumn{1}{r|}{0.005} & 439.4$\pm$11.4 & \textbf{383.3$\pm$0.9} & 253.6$\pm$11 & \textbf{215.0 $\pm$ 0.3} \\
\multicolumn{1}{r|}{0.01} & 389.9$\pm$1.8 & \textbf{368.9$\pm$0.5} & 222.7$\pm$0.9 & \textbf{212.8 $\pm$ 0.1} \\
\multicolumn{1}{r|}{0.05} & \textbf{392.1$\pm$0.8} & 421.1$\pm$0.5 & \textbf{222.4$\pm$0.2} & 227.2 $\pm$ 0.1 \\
\multicolumn{1}{r|}{0.1} & \textbf{418.7$\pm$5.9} & 461.9$\pm$0.3 & \textbf{227.6$\pm$0.5} & 242.4 $\pm$ 0.4 \\ \bottomrule
\label{tab:lr_study}
\end{tabular}
\end{table}

\clearpage
\section{Dataset Statistics}\label{app:data}
Our dataset pre-processing for \emph{german}, \textit{svmguide3}, \textit{magic04}, \textit{a8a}, \textit{SUSY} and \textit{HIGGS}  follows exactly the steps from~\citet{agarwal2023auxdrop}. The data is publicly available for download online\footnote{Available here:~\url{https://github.com/Rohit102497/Aux-Drop/tree/main/Code/Datasets}}.
For CIFAR-10 we use the standard training set that consists of 50,000 images. Image pixel intensities are normalized to fall within $[0,1]$ range.
For I-MNIST there used to exist an online repository of the dataset. However, this repository appears to have been removed. As a result we rebuild the dataset using the source code that originally generated the data from L. Bottou\footnote{Available here: \url{https://leon.bottou.org/projects/infimnist}}. We use the default configuration settings provided in the repository.
We provide summary dataset statistics in Tab.~\ref{tab:data}. The use of diverse dataset sizes aims to illustrate the strength of our approach in settings with low data (one thousand), moderate amount (tens of thousands) and large data (millions).
\begin{table}[h]
\centering
\caption{Dataset statistics for the data used in our experiments.}
\begin{tabular}{@{}lccc@{}}
\toprule
\textbf{Dataset} & \textbf{Size} & \textbf{Feature Size} & \textbf{Task} \\ \midrule
german & 1000 & 24 & Classification \\
svmguide3 & 1243 & 21 & Classification \\
magic04 & 19020 & 10 & Classification \\
a8a & 32561 & 123 & Classification \\
CIFAR-10 & 50000 & 32$\times$32$\times$3 & Classification \\
I-MNIST & 1000000 & 28$\times$28 & Classification \\
SUSY & 1000000 & 8 & Classification \\
HIGGS & 1000000 & 21 & Classification \\ \bottomrule
\end{tabular}\label{tab:data}
\end{table}

\section{Score Sum Validation Implementation Details}\label{app:details}

Due to space constraints in the main paper we limited discussion of the details of the baseline designs for this ablation study. Here we add concrete details for each baseline. The results for the baselines described below, comparing them against our method, appear in Table 4 of the original paper.

Denote each learner output as $f_i$. Note that for this ablation study we fix the learner pool and the architectures of the learner models to be the same as MODL in Tables 1, 2. Below, we provide the detailed description of each baseline.

\textbf{Mixture of experts (MoE)} model learns a gating mechanism to combine model scores $f_i$. Note that standard MoE practice is to combine logit probabilities rather than latent scores so we apply a softmax function to each learner to produce a valid probability distribution. The gating network ${G}$, parameterized by $\Theta$ is a 2 layer softmax network, yielding the output ${G}(x; \Theta)$:

\begin{align}
F_{\text{MoE}}({x}; {\Theta}, \{{W}_i\}_{i=1}^N) &= \sum_{i=1}^{N} {G}({x}; {\Theta})_i [\text{softmax}(f_i)],\\
{G}({x}; {\Theta})_i &= \text{softmax}(g({x}; {\Theta}))_i = \frac{\exp(g({x}; {\Theta})_i)}{\sum_{j=1}^{N} \exp(g({x}; {\Theta})_j)},
\end{align}

where $g(\cdot)$ is the value at the penultimate layer of the neural network parameterizing $G(x, \Theta)$. Note that this approach is a generic Dense MoE that has been commonly applied in the literature ("Mixture of experts: a literature survey", Masoudnia and Ebrahimpour, 2014). Note that backpropagation steps are taken from the output of $F_{\text{MoE}}$, and therefore the models are trained jointly here.

\textbf{Ensemble} (Deep Ensembles - \citet{lakshminarayanan2017ensembles}) model combines the predicted class probabilities from each model as follows:

\begin{align}
F_{\text{DE}}({x}) = \frac{1}{N}\sum_{i=1}^{N}  \text{softmax}(f_i),
\end{align}

where the models are trained independently (backpropagation passes are separate).
    
\textbf{Greedily Weighing} is the online weighted ensemble method that weighs models proportionally to their online accuracy within a sliding window. Concretely, this takes the form:

\begin{align}
F_{\text{Greedy}}({x}) &= \sum_{i=1}^{N}  \left(\text{softmax}([\lambda_{t}])_i\right) \, \text{softmax}(f_i),\\
[\lambda_{t}]_i &= \sum_{j = t-K}^{t-1} \frac{\mathbb{I}[f_i(x_j) = y_j]}{K},
\end{align}
    
where $\lambda_{t}$ is a vector of length $N$ with the running accuracy of each of the model within the sliding window of size $K$, $[\lambda_{t}]_i$ is the $i$-th index of the vector (and contains the running accuracy of $i$-th model) and $\mathbb{I}[\cdot]$ is an indicator function. In our experiments we chose $K=100$ as a reasonable tradeoff to achieve responsiveness to model performance change but also robustness to temporary fluctuations in model performance as the models converge. If $K$ is set too low, e.g., below 10, then the method can become unstable. Again, the ensemble methods here are trained independently  (backpropagation passes are separate).
    
\textbf{Multiplication} baseline interprets each model output (after it is projected to a valid probability distribution) as an independent estimate of the class probability. Hence, in order to obtain a distribution over all models, it multiplies the constituent distributions and re-normalizes. For numerical stability this is done via logarithm addition, which also has a clear interpretaion as the sum ensemble in the logit domain:

\begin{align}
F_{\text{Mult}}({x}) &= \text{softmax}\left(\sum_{i=1}^{N}  \log \text{softmax}(f_i)\right).
\end{align}

Here, models are entangled via the softmax, and the backpropagation pass operates on the combined output and thus trains the models jointly.

\section{Hyperparameters and Computational Resources}\label{app:hparams}
\subsection{Hyperparameters}
\textbf{Aux-Drop (ODL)}: We used the official repository of Aux-Drop (ODL)\footnote{Available here:~\url{https://github.com/Rohit102497/Aux-Drop}} with the tuned hyperparameters from the original paper by~\citet{agarwal2023auxdrop} without making any changes. Namely, for the small datasets in our experiments we run 6-layer networks with learning rates 0.1 for german and svmguide3 and 0.01 for magic04 and a8a. For the large datasets, we used 11 layer networks with learning rate equal to 0.05. 
For image datasets we run Aux-Drop using 20 learning rates between $[10e-6, 10e-2]$ and reported the results of the best setup ($5e-4$).
For all datasets we put the AuxLayer in the third layer of the network. 
For \emph{german}, \textit{svmguide3}, \textit{magic04}, \textit{SUSY} and \textit{HIGGS} the capacity of the hidden layer is 50 and in the capacity of AuxLayer is 100. For \textit{a8a} the AuxLayer has 400 units. For CIFAR-10 and I-MNIST we use multiple hidden unit layer capacities $\{ 100, 250, 500 \}$ and report the results on the best one (250).

For all experiments we used the recommended layer size and AuxLayer dimension and position in the network. Additionally, the ODL discount rate was set to $\beta=0.99$ and the smoothing rate was set to $s=0.2$.

\textbf{MODL}: We used the exact same architecture in all  experiments. Specifically, the online logistic regression learner is implemented identically for all datasets and the MLP has 3 hidden layers with 250 neurons. The set learner consists of 6 blocks with 3 layers per block. The width of each layer is set to 128 neurons. The overall performance is stable for various layer widths. For the MLP and the set learner we observe that among the capacities of $\{ 100, 250, 500 \}$ neurons per layer the best performance is given by 250 units but the difference is not very large. For the set learner we validated that the number of blocks in the encoder and decoder (candidate values $\{2,3,4,5,6,7,8\}$) and the number of layers per block (candidate values $\{ 2,3,4,5 \}$) do not have a huge performance impact, though deeper architectures perform better for larger datasets and more narrow ones have an edge in smaller datasets. Our selected setup for all datasets has good all around performance.

\begin{table}[ht]
\tiny
\centering
\caption{Hyperparameters for MODL in all experiments.}
\begin{tabular}{@{}lcccccc@{}}
\toprule
\textbf{Dataset} &
  \textbf{Learning Rate} &
  \textbf{Layers MLP} &
  \textbf{MLP Layer Width} &
  \multicolumn{1}{l}{\textbf{Blocks Set Learner}} &
  \multicolumn{1}{l}{\textbf{Layers Block}} &
  \multicolumn{1}{l}{\textbf{Block Layer Width}} \\ \midrule
german    & 0.01    & 3 & 250 & 6 & 3 & 250 \\
svmguide3 & 0.01    & 3 & 250 & 6 & 3 & 250 \\
magic04   & 0.001   & 3 & 250 & 6 & 3 & 250 \\
a8a       & 0.001   & 3 & 250 & 6 & 3 & 250 \\
CIFAR-10  & 0.00005 & 3 & 250 & 6 & 3 & 250 \\
I-MNIST   & 0.00005 & 3 & 250 & 6 & 3 & 250 \\
SUSY      & 0.005   & 3 & 250 & 6 & 3 & 250 \\
HIGGS     & 0.005   & 3 & 250 & 6 & 3 & 250 \\ \bottomrule
\end{tabular}
\end{table}

\subsection{Computational Resources}
We run our experiments on a HP DL-580 Gen 7 server equipped with Intel Xeon E7-4870 2.40GHz 10-Core 30MB LGA1567 CPUs. In total we used 160 CPUs for our experiments. Note that using GPUs in the online learning setting is not effective as the batch size is 1. Therefore, given that our task is purely sequential we opt for a CPU server that is actually faster than a GPU server given the nature of the online learning task.

\section{Bayesian Filters for Online Regression} %
In this section we present the key theoretical result that underpins our online logistic regression learner. This result stems from De Finetti's representation theorem. This key result allows us to derive recursive algorithms that update prior-posterior parameters without storing observations, an essential element in any dataset-memoryless online algorithm. We then review a standard result for online linear regression which we then generalize via linearization and a Normal approximation to obtain the online logistic regression learner we use in our algorithm.

\subsection{De Finetti Representation Theorem}

For $n\geq1$ the de Finetti representation for the joint mass or density of exchangeable discrete random
variables $Y_1,\dots,Y_n$ is given by
$$p_{Y_1,\dots,Y_n}(y_1,\dots,y_n)=\int_\Theta \prod_{i=1}^n f_{Y}(y_i; \theta) \pi_0(d\theta)$$
where $p_Y(y; \theta)$ is a mass function in $y$, and $\theta$ is a parameter lying in a space $\Theta \in \mathbb{R}^p$, for some (prior)
distribution $\pi_0(d\theta)$ defined on $\theta$, where we may interpret $\Theta$ as the smallest set such that $\int_\Theta \pi_0(d\theta)=1.$
Using basic properties of probability we may derive an expression for the posterior predictive distribution for $Y_{n+1}$ (another
element of the infinitely exchangeable sequence) conditional on $Y_1 = y_1, \dots , Y_n = y_n$. This defines the posterior density, $\pi_n(d\theta)$, as an updated version of $\pi_0(d\theta)$.
We may re-write the posterior predictive distribution using the definition of conditional probability as
\begin{align}
    p_n(y_{n+1}) = \frac{p_0(y_1, \cdots, y_n, y_{n+1})}{p_0(y_1, \cdots, y_n)} = \int f_Y(y_{n+1}; \theta)\pi_n(\theta) \, d\theta
\end{align}
where 
\begin{align}
    \pi_n(\theta) = \frac{1}{p_0(y_1, \cdots, y_n)}\prod_{i=1}^{n}f(y_i; \theta)\pi_0(\theta),
\end{align}
is the posterior. Note that this is the same form as the De Finetti representation with an updated version of $\pi_0$.

\noindent This result is important as it shows that the posterior predictive is the same whether we consider the initial prior and the likelihood of a full batch on $n$ observations or if we use an iterative update rule to modify the prior at each time step. Thus, we may simply update the prior for the $n{+}1$-th observation to be the posterior after the $n$-th observation without the need to maintain any of the previous observations in memory. This allows us to derive recursive algorithms that update prior-posterior parameters without storing observations. Essentially, as long as we maintain a statistic that is sufficient in the Bayesian sense, the posterior distribution depends on the data
only through the observed value of the statistic and the data may be discarded.

\subsection{Bayesian Linear Regression}
For a dataset $\dataset=\{ (t_1,y_1), \dots, (y_N, t_N) \}$, consider a classic regression model of the form:
\begin{align}
    y_k = \vec x_k \vec \theta + \varepsilon_k,
\end{align}
where $y_k$ is a scalar outcome, $\vec \theta\in\Re^{d\times 1}$ is the parameter of interest, $\vec x_k\in\Re^{1\times d}$ is the feature vector and $\varepsilon_k\sim \mathcal{N}(0, \sigma^2)$ is normally distributed noise. To simplify the notation we may stack all feature vectors in the designer matrix $\vec X\in\Re^{N\times d}$ and the corresponding outcomes in a vector $\vec y_k \in\Re^{N}$. 

\bigskip

\noindent Though the problem can be solved in a purely least squares fashion or equivalently via a maximum-likelihood approach, we employ a standard Bayesian formulation. Assigning the prior,
\begin{align}
    \pi_0(\vec \theta) = \mathcal{N}(\vec \theta \, \lvert \, \vec m_0, \vec P_0)\,,
\end{align}
for some $\vec m_0\in\Re^{d}, \vec P_0\in\Re^{d\times d}$. and modeling $p(\vec y_k \, \lvert \, \vec \theta)$ as 
\begin{align}
    p(y_k \, \lvert \, \vec \theta) = \mathcal{N}(y_k \, \lvert \, \vec x_k\vec \theta, \sigma^2),
\end{align}
we may then apply Bayes' rule to recover the posterior distribution over the parameter:
\begin{align}
    p(\theta \, \lvert \, y_{1:N}) &\propto p(\theta) \prod_{k=1}^N p(y_k \, \lvert \, \theta) = \mathcal{N}(\vec \theta \, \lvert \, \vec m_0, \vec P_0) \prod_{k=1}^N \mathcal{N}(y_k \, \lvert \, \vec x_k\vec \theta, \sigma^2)\\
    &= \mathcal{N}(\vec \theta \, \lvert \, \vec m_N, \vec P_N)\,.
\end{align}
The last equality follows from the well known result that normal priors and normal likelihoods yield normal posteriors. In fact, it can be shown that
\begin{align}
    \vec m_N &= \left[ \vec P_0^{-1} + \frac{1}{\sigma^2} \vec X^\top \vec X\right ]^{-1}\left[ \frac{1}{\sigma^2} \vec X^\top \vec y + \vec P^{-1}_0\vec m_0 \right],\\
    \vec P_N &= \left[ \vec P_0^{-1} + \frac{1}{\sigma^2} \vec X^\top \vec X\right]^{-1}.
\end{align}

\subsection{Online Bayesian Linear Regression}

A disadvantage of full batch regression is that it requires keeping a large matrix $\left[ \vec P_0^{-1} + \frac{1}{\sigma^2} \vec X^\top \vec X\right]^{-1}$ in memory and inverting it. To address these concerns, we can use an efficient and recursive version of the algorithm. From the recursive de Finetti represention result we presented earlier, we may alternatively calculate a posterior at step $n$-$1$ and treat it as a prior for step $n$.
\begin{align} \label{eqn:posterior_parameter_model}
    p(\theta \, \lvert \, y_{1:n}) &\propto p(\theta \, \lvert \, y_{1:n-1})  p(y_n \, \lvert \, \theta) = \mathcal{N}(\vec \theta \, \lvert \, \vec m_{n-1}, \vec P_{n-1}) \mathcal{N}(y_k \, \lvert \, \vec x_k\vec \theta, \sigma^2)\\
    &= \mathcal{N}(\vec \theta \, \lvert \, \vec m_n, \vec P_n),
\end{align}
where
\begin{align} \label{eqn:mean_and_variance}
    \vec m_n &= \left[ \vec P_{n-1}^{-1} + \frac{1}{\sigma^2} \vec X_n^\top \vec X_n\right ]^{-1}\left[ \frac{1}{\sigma^2} \vec X_n^\top \vec y_n + \vec P_{n-1}^{-1}\vec m_{n-1} \right],\\
    \vec P_n &= \left[ \vec P_{n-1}^{-1} + \frac{1}{\sigma^2} \vec X_n^\top \vec X_n\right]^{-1}.
\end{align}
We use the notation, $\vec X_n$ and $\vec y_n$ to indicate the $n$-th rows of $\vec X$, $\vec y$, respectively.

\noindent The matrix inversion lemma\footnote{\url{https://en.wikipedia.org/wiki/Woodbury_matrix_identity}}, also known as the Woodbury matrix identity, states that:
\begin{align} \label{eqn:application_of_matrix_inversion_lemma}
    \vec P_n = \left[ \vec P_{n-1}^{-1} + \frac{1}{\sigma^2} \vec X_n^\top \vec X_n\right]^{-1} = \vec P_{n-1} - \vec P_{n-1} \vec X_n^\top \left [ \vec X_n \vec P_{n-1} \vec X_n^\top + \sigma^2\right]^{-1} \vec X_n \vec P_{n-1}.
\end{align}
Note that the inverse term in the expression on the right hand side is a scalar and thus easy to compute.

\noindent The recursive set of equations takes the form:
\begin{align}
    S_n &= \vec X_n \vec P_{n-1} \vec X_n^\top + \sigma^2,\\
    \vec K_n &=  \vec P_{n-1} \vec X_n^\top S_n^{-1},\\
    \vec m_n &= \vec m_{n-1} + \vec K_n [\vec y_n - \vec X_n \vec m_{n-1}],\\
    \vec P_{n} &=  \vec P_{n-1} - \vec K_n S_n \vec K_n^\top.
\end{align}

\section{Logistic Regression}

Consider a Generalized Linear Model (GLM) with link function $g(\cdot)=\text{logit}^{-1}(\cdot)$.
For binary classification we define $p(y \, \lvert \, x) = \mathbb{E}_{Y|X}[y \,\lvert\, \vec x] = g^{-1}(\vec x \theta) = \sigma(\vec x \theta) = \mu$, $\sigma(\cdot)$ is the sigmoid function.
The problem of finding the optimal parameter can be solved by maximizing the (log) likelihood numerically as it is convex. More precisely the derivative of the log-likelihood and the Hessian can be shown to be
\begin{align}
    &\dot{\ell}_n(\theta) = \vec X^\top(\vec y -  \boldsymbol{\mu}), \\
    &\ddot{\ell}_n(\theta) = -\vec X^\top \vec D \vec X,
\end{align}
where $\boldsymbol{\mu}=[\sigma(\vec X_1\theta), \dots, \sigma(\vec X_n\theta)]$ is a vector containing the model prediction for each input, $\vec D$ is a diagonal matrix with the diagonal entries $[\sigma(\vec X_1\theta)(1-\sigma(\vec X_1\theta)), \dots, \sigma(\vec X_n\theta)(1-\sigma(\vec X_n\theta))].$ Since this is a convex problem, it has a unique solution that can be recovered via standard convex optimization techniques.

\section{Online Logistic Regression Derivation}\label{app:derivation}

\subsection{Proof of Proposition~\ref{prop:glm}}

Consider an input $\vec z \in \Re^n$ and output $\vec y \in \Re^m$, that are related via an invertible function $h(\cdot): \Re^{n} \to \Re^m$ 
\begin{align}
    &\vec z \sim \mathcal{N}(\vec z \, \lvert \, \vec m, \vec P)\\
    &\vec y = h(\vec z) + \varepsilon_t,
\end{align}
where we model the output as receiving additive noise $\varepsilon\sim\mathcal{N}(\varepsilon_t \, | \, 0, \boldsymbol{\Sigma}_t)$ that is independent of the input. By standard random variable transformation theory we know that
$p(\vec y) =  \mathcal{N}(h^{-1}(\vec y) \, \lvert \, \vec m, \vec P) |\mathcal{J}(\vec y)|^{-1}$, where $|\mathcal{J}(\vec y)|$ is the Jacobian of $h$ evaluated at $\vec y$.

A challenge in analyzing this model is that it is non-linear.
A standard technique to approximate $p(\vec y)$ is to apply a first order Taylor approximation and linearize the function locally around the mean of the normal $\mathcal{N}(\vec z \, \lvert \, \vec m, \vec P)$ with some random perturbation $\delta\vec z \sim \mathcal{N}(\delta\vec z \,|\,0,\mathbf{P})$:
\begin{align}
    h(\vec z) = g(\vec m + \delta\vec z) \approx g(\vec m) + \mathcal{J}(\vec m)\delta\vec z = \widehat{h}(\vec z).
\end{align}

\noindent Firstly, note that $\mathbb{E}[h(\vec z)] \approx \mathbb{E}[h(\vec m)]  + \mathcal{J(\vec m)}\mathbb{E}[\delta \vec z] = \mathbb{E}[h(\vec m)] $.
Similarly, for the covariance matrix it follows that cov$(h(\vec x)) = \mathbb{E}\left[ (h(\vec z) - \mathbb{E}[h(\vec z)]) (h(\vec z) - \mathbb{E}[h(\vec z)])^\top \right]\approx \mathcal{J}(\vec m) \vec P \mathcal{J}(\vec m)^\top.$ To show this in detail consider:
\begin{align}\label{eq:covariance}
    \text{cov}(h(\vec z)) &= \mathbb{E}\left[ (h(\vec z) - \mathbb{E}[h(\vec z)]) (h(\vec z) - \mathbb{E}[h(\vec z)])^\top \right]\\
    &\approx \mathbb{E}\left[ (h(\vec z) - h(\vec m)) (h(\vec z) - h(\vec m))^\top \right]\\
    &=\mathbb{E}\left[ (h(\vec m) + \mathcal{J}(\vec m)\delta \vec z - h(\vec m)) (h(\vec m) + \mathcal{J}(\vec m)\delta \vec z - h(\vec m))^\top \right]\\
    &=\mathbb{E}\left[ (\mathcal{J}(\vec m)\delta \vec z) (\mathcal{J}(\vec m)\delta \vec z) ^\top \right]\\
    &= \mathcal{J}(\vec m)\mathbb{E}\left[(\delta \vec z)(\delta \vec z) ^\top \right] \mathcal{J}(\vec m)^\top\\
    &=\mathcal{J}(\vec m) \vec P \mathcal{J}(\vec m)^\top.
\end{align}
\noindent Combining the previous results we obtain an approximate joint distribution of $\vec z, \vec h(\vec z)$:
\begin{equation}
    p\left(\begin{bmatrix}
    \vec z \\
    \vec y
\end{bmatrix}\right)
= \mathcal{N}\left(
\begin{bmatrix}
    \vec m \\
    h(\vec m)
\end{bmatrix}, 
\begin{bmatrix}
    \vec P & [\vec P\mathcal{J}(\vec m)]^\top  \\
    \vec P \mathcal{J}(\vec m) &  \boldsymbol{\Sigma}_t+ \mathcal{J}(\vec m)\vec P \mathcal{J}^\top(\vec m)
\end{bmatrix}
\right)\,.
\end{equation}

In the case of input features that are multiplied by the regression parameters, we have $h(\vec z)\eqdef h(\vec x \theta)$
where the input feature vector $\vec x$ is multiplied by the weights vector $\theta$, \emph{i.e.}, $\vec z \eqdef \vec x \theta$. This modifies the joint distribution we just derived:
\begin{align}
    &\vec \theta \sim \mathcal{N}(\vec \theta \, \lvert \, \vec m, \vec P)\\
    &\vec y = h(\vec x \theta) + \varepsilon,
\end{align}
\begin{equation}
    p\left(\begin{bmatrix}
    \vec \theta \\
    \vec y
\end{bmatrix}\right)
= \mathcal{N}\left(
\begin{bmatrix}
    \vec m \\
    h(\vec m \theta)
\end{bmatrix}, 
\begin{bmatrix}
    \vec P & [\vec P\vec x\mathcal{J}(\vec m)]^\top  \\
    \vec P \vec x \mathcal{J}(\vec m) & \boldsymbol{\Sigma}_t + \mathcal{J}(\vec m)\vec x\vec P\vec x^\top\mathcal{J}^\top(\vec m)
\end{bmatrix}
\right)\,.
\end{equation}

Then, from the joint distribution $p(\theta, \vec y)$ we may obtain the conditional $p(\theta \,|\, \vec y)$, which has a well known closed form for jointly normal random variables, 
\begin{align}
    p(\vec \theta \, | \, \vec y = y_k) &= \mathcal{N}(\vec \theta \, | (\vec m', \,\vec P'), \quad \text{where}\\
    \vec m'&= \vec m + [\vec P \vec x \mathcal{J}(\vec m)]^\top [\boldsymbol{\Sigma}_k + \mathcal{J}(\vec m)\vec x\vec P\vec x^\top\mathcal{J}^\top(\vec m)]^{-1} (\vec y_k - h(\vec m \theta))\\
    \vec P' &= \vec P - [\vec P \vec x \mathcal{J}(\vec m)]^\top [\boldsymbol{\Sigma}_k + \mathcal{J}(\vec m)\vec x\vec P\vec x^\top\mathcal{J}^\top(\vec m)]^{-1} [\vec P \vec x \mathcal{J}(\vec m)]
\end{align}
Thus we obtain a normal posterior over the weights.
But observe that this is just another recursive normal prior --- normal likelihood model. Thus we may obtain recursive form updates similar to those we derived for the linear regression earlier.
\begin{align} %
    p(\theta \, \lvert \, y_{1:n}) &\propto p(\theta \, \lvert \, y_{1:n-1})  p(y_n \, \lvert \, \theta) = \mathcal{N}(\vec \theta \, \lvert \, \vec m_{n-1}, \vec P_{n-1}) \mathcal{N}(y_n \, \lvert \, h(\vec x_n\vec \theta), \sigma^2)\\
    &= \mathcal{N}(\vec \theta \, \lvert \, \vec m_n, \vec P_n),
\end{align}
where
\begin{align}
    \vec m_n &= \vec m_{n-1} + \frac{[\vec P_{n-1} \vec x_n^\top \mathcal{J}(\vec m_{n-1})]^\top [\vec y_n - h\left(\vec x_n \vec m_{n-1}\right)]}{\boldsymbol{\Sigma}_n + \mathcal{J}(\vec m_{n-1})\vec x_n\vec P_{n-1}\vec x_n^\top\mathcal{J}^\top(\vec m_{n-1})},\\
    \vec P_{n} &=  \vec P_{n-1} - \frac{\mathcal{J}^\top(\vec m_{n-1})\vec P_{n-1}\vec x_n^\top \vec x_n \vec P_{n-1}^\top\mathcal{J}(\vec m_{n-1})}{\boldsymbol{\Sigma}_n + \mathcal{J}(\vec m_{n-1})\vec x_n\vec P_{n-1}\vec x_n^\top\mathcal{J}^\top(\vec m_{n-1})}.
\end{align}

\paragraph{Online Binary Logistic Regression} Concretely, for a binary logistic regression $h(\cdot)\eqdef \text{logit}^{-1}(\vec x \theta) = \sigma(\vec x \theta)$ we can
calculate the Jacobian to obtain a closed form update.
\begin{align} \label{eqn:posterior_parameter_model_2}
    p(\theta \, \lvert \, y_{1:n}) &= \mathcal{N}(\vec \theta \, \lvert \, \vec m_n, \vec P_n),
\end{align}
\begin{align}
    \vec m_n &= \vec m_{n-1} + \frac{\vec P_{n-1} \left[\left(1-\vec \sigma(\vec x_n \vec m_{n-1})\right) \sigma\left(\vec x_n \vec m_{n-1}\right)\right]\vec x_n^\top [\vec y_n - \sigma\left(\vec x_n \vec m_{n-1}\right)]}{\boldsymbol{\Sigma}_n + \vec P_{n-1} \left[\left(1-\vec \sigma(\vec x_n \vec m_{n-1})\right) \sigma\left(\vec x_n \vec m_{n-1}\right)\right]^2},\\
    \vec P_{n} &=  \vec P_{n-1} - \frac{\vec P_{n-1}\vec x_n^\top \vec x_n \vec P_{n-1}^\top \left[\left(1-\vec \sigma(\vec x_n \vec m_{n-1})\right) \sigma\left(\vec x_n \vec m_{n-1}\right)\right]^2}{\boldsymbol{\Sigma}_n + \vec P_{n-1} \left[\left(1-\vec \sigma(\vec x_n \vec m_{n-1})\right) \sigma\left(\vec x_n \vec m_{n-1}\right)\right]^2}.
\end{align}
This yields the following update equations:
\begin{align}
    S_n &= \boldsymbol{\Sigma}_n + \mathcal{J}(\vec m_{n-1})\vec P_{n-1} \mathcal{J}^\top(\vec m_{n-1}),\\
    \vec K_n &=  \vec P_{n-1} \vec x_n^\top \mathcal{J}(\vec m_{n-1})^\top S_n^{-1},\\
    \vec m_n &= \vec m_{n-1} + \vec K_n [\vec y_n - h\left(\vec x_n \vec m_{n-1}\right)],\\
    \vec P_{n} &=  \vec P_{n-1} - \vec K_n S_n \vec K_n^\top.
\end{align}
\begin{align}
    S_n &= \boldsymbol{\Sigma}_n + \vec P_{n-1} \left[\left(1-\vec \sigma(X_n \vec m_{n-1})\right) \sigma\left(\vec X_n \vec m_{n-1}\right)\right]^2,\\
    \vec K_n &=  \vec P_{n-1} \vec x_n^\top\left(1-\vec \sigma(X_n \vec m_{n-1})\right)\sigma\left(\vec X_n \vec m_{n-1}\right) S_n^{-1},\\
    \vec m_n &= \vec m_{n-1} + \vec K_n [\vec y_n - \sigma\left(\vec X_n \vec m_{n-1}\right)],\label{eq:update}\\
    \vec P_{n} &=  \vec P_{n-1} - \vec K_n S_n \vec K_n^\top.
\end{align}
To recover the result in the main paper we may choose to model the process noise $\boldsymbol{\Sigma}_n \eqdef \vec x_n\vec P_{n-1}\vec x_n$. This provides a regularization effect by not allowing the update steps to become very large when $\vec x_n$ has a large magnitude.

\paragraph{Online Multinomial Logistic Regression}

In the multinomial regression case, the features have the same dimensions, but the observations are a one hot vector $\vec y\in\mathbb{R}^K$. Here $K$ is the number of classes. Each of the $K$ classes has an associated parameter vector $\vec \theta^{(k)} \in \mathbb{R}^{D}$, $k\in{1, \dots, K}$. Stacking these vectors yields a parameter matrix $\Theta \in \mathbb{R}^{D \times K}$ with element $\Theta^{(k)}_d$ denoting the weight of input feature at index $d$ for the $k$-th class.

The function that maps the input vector to the prediction is the softmax function $\hat{p}_l = \frac{\exp( {\Theta^{(l)}}^T \mathbf{x})}{1 + \sum^K_{j=1} \exp({\Theta^{(j)}}^T \mathbf{x})}$, leading to log likelihood:  
\begin{equation}
    \mathcal{L}(\Theta) = \left( \sum^K_{k=1} \vec y_k {\Theta^{(k)}}^T \mathbf{x} \right) - \ln \left(1 + \sum^K_{j=1} \exp( {\Theta^{(j)}}^T \mathbf{x})  \right).
\end{equation}
One can show that the gradient of this function evaluates to the following Kronecker product~\citep{bhning1992multinomialLR}:
\begin{align}
\nabla_{\Theta} \mathcal{L}(\Theta) =
\begin{bmatrix}
(\mathbf{y} - \hat{\mathbf{p}})_1 \mathbf{x} \\
(\mathbf{y} - \hat{\mathbf{p}})_2 \mathbf{x} \\
\vdots \\
(\mathbf{y} - \hat{\mathbf{p}})_K \mathbf{x} \\
\end{bmatrix}
= (\mathbf{y} - \hat{\mathbf{p}}) \otimes \mathbf{x},
\end{align}
which has the same dimensions as the weight matrix $\Theta$. Thus, in the multinomial case we may still apply the closed form updates derived previously by setting $\mathcal{J}(\Theta) = \nabla_{\Theta} \mathcal{L}(\Theta)$. This complicates the updates as the parameter is now a matrix but the update complexity remains constant in the size of the parameter and we never have to perform any matrix inversions.

\subsection{Relation to other Online Logistic Regression approaches}
There are two main other approaches to online logistic regression that are relevant to our method. First, FOLKLORE by~\citet{agarwal2021folklore} proposes an iterative optimization scheme where logistic regression parameters can be learned by iteratively solving an optimization problem. While this approach can work well, it lacks the closed form updates that are needed to improve efficiency. An approach with closed form updates is offered in the work of~\citet{vilmarest2021kalman}. These updates are quite similar to ours, and if we remove the regularization from our approach can be made equivalent. However,~\citet{vilmarest2021kalman} arrive at these updates using Kalman filter theory to derive general results about exponential family models whereas our derivation uses much more basic tools from probability theory. We believe that our approach is intuitive and the derivation makes the linearization assumptions more obvious to the reader rather than automatic application of Kalman filters.

\subsection{Toy Experiment}\label{app:toy_exp}
\begin{figure*}[t]
    \centering
    \includegraphics[width=0.43\textwidth, height=0.4\textwidth]{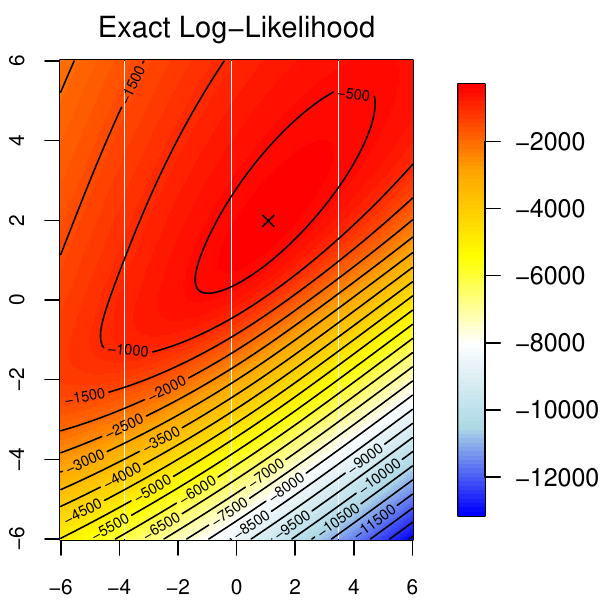}
    \includegraphics[width=0.43\textwidth, height=0.4\textwidth]{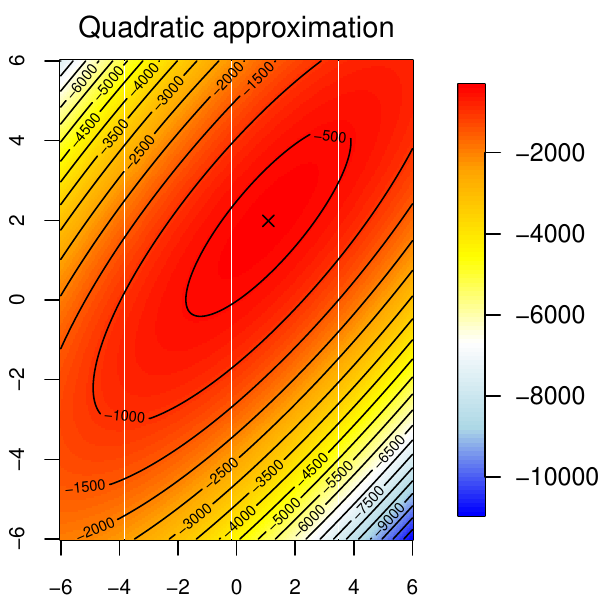}
    \caption{Exact log-likelihood vs. our proposed quadratic approximation. We see that close to the data-generating parameters (marked as x) both our approximation and the exact log-likelihood function agree. See~\Cref{app:toy_exp} for toy dataset experiment details.}
    \label{fig:quad_approx}
\end{figure*}

In this section we present a toy experiment that analyzes the online logistic regression. We consider the following toy data generating process for a binary logistic regression model with two data generating parameters $\theta^*$:
\begin{align}
    \theta^* &= (1, 2)\\
    \vec X &\sim\mathcal{N}(\vec X \,|\,-1.5,1)\\
    \vec Y &= \begin{cases}
        0, \quad \text{with probability}\,\,\, \frac{1}{1 + \text{exp}(\vec x \cdot \theta)}\\
        1, \quad \text{with probability}\,\,\, 1-\frac{1}{1 + \text{exp}(\vec x \cdot \theta)}
    \end{cases}
\end{align}
We generate 100 pairs by drawing independently 100 times from $X$ and generating $Y$ given each sample. Note that $\vec X\in\mathbb{R}^2$ is augmented with a constant value 1 to handle the intercept parameter. We observe that the log-likelihood shown in Fig.~\ref{fig:quad_approx} matches well with the quadratic log likelihood approximation around the data generating values.

\newpage

\section{Architectural Details}

\begin{figure}[h]
    \centering
    \includegraphics{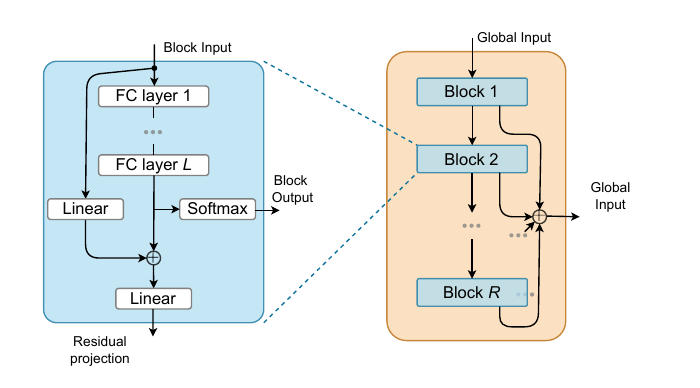}
    \caption{Visualization of our set learning decoder.}
    \label{fig:block}
\end{figure}
\textbf{Slow learner with set inputs.} Rather than masking missing features with a zero or using deterministic dropout, as is done in~\citep{agarwal2023auxdrop}, we treat the input as a set that excludes any missing features.
Recall that the data generating process produces a sequence of triplets $\process=\{(\vec{z}_1, \vec{y}_1, \mask_1),\dots, (\vec{z}_T, \vec{y}_T, \mask_T)\}$. 
Then, the set of input features $\mathcal{X}_t$ can be expressed as:
\begin{align}%
    \mathcal{X}_t &= \left\{\vec{z}_{t,j} : \mask_{t,j} = 1\right\}, \quad 
    \mathcal{I}_t = \left\{j : \mask_{t,j} = 1\right\}, \quad
    \dataset=\left[(\mathcal{X}_1, \mathcal{I}_1, \vec{y}_1), \dots, (\mathcal{X}_T,  \mathcal{I}_T, \vec{y}_T)\right].
\end{align}
The size of the input feature set $\mathcal{X}_t$ is time varying. To allow the model to determine which inputs are available, it is necessary to pass a set of feature IDs in an index set $\mathcal{I}_t$. Our proposed set learning module follows closely the ProtoRes architecture (please refer to~\citet{oreshkin2022protores} for details). It takes the index set $\mathcal{I}_t$ and maps each of its active index positions to a continuous representation to create feature ID embeddings. It then concatenates each ID embedding to the corresponding feature value of $\mathcal {X}_t$. These feature values and ID embedding pairs are aggregated and summed to produce fixed dimensional vector representations that
we denote as $\vec x_0$.
The main components of our proposed set learning module are 
blocks, each consisting of $L$ fully connected (FC) layers. Residual skip connections are included in the architecture so that blocks can be bypassed.
An input set $\mathcal X$ is mapped to $\vec x_0 = \textsc{EMB}(\mathcal{X})$. The overall structure of the module at block $r\in\{1, \dots,R\}$ is:
\begin{align}
    \vec h_{r,1} &= \textsc{FC}_{r,1}(\vec x_{r-1}), \quad \dots, \quad \vec h_{r,L} = \textsc{FC}_{r,L}(\vec h_{r,L-1}),\\
    \vec x_r &= \textsc{ReLU}(\vec W_r\vec x_{r-1} + \vec h_{r-1,L})\,,  \quad \hat{\vec y}_{r} 
    = \hat{\vec y}_{r-1} + \vec Q_L \vec h_{r,L},
\end{align}
where $\vec W_r, \vec Q_L$ are learnable matrices. We connect $R$ blocks sequentially to obtain global output $\hat{\vec y}_R$.

\newpage

\section{Fast Online Deep Learning - Additional Details}

\subsection{Proof of Proposition~\ref{prop:backprop}}\label{app:backprop}
The proof is an inductive argument.
First, we analyze a base case for a $L=3$ layer network for a proof sketch.

\textbf{BASE CASE: Prove the proposition holds for} $L=3$.

\begin{figure}[ht!]
    \centering\includegraphics[height=9cm]{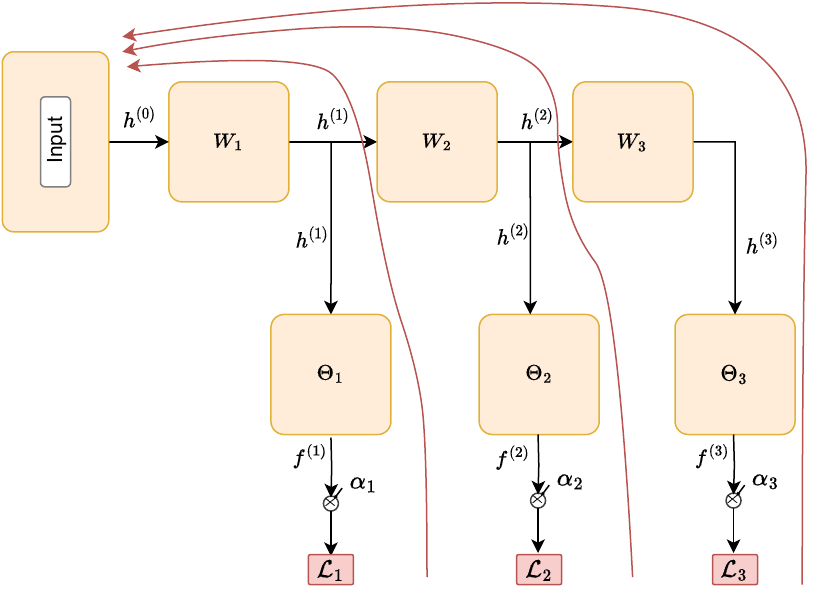} 
    \caption{ODL hedge backpropagation. Red lines indicate individual backpropagation calculations.}
    \label{fig:odl_backprop}
\end{figure}

Recall the variable definitions:
\begin{align}
\nonumber  f^{(l)}   &= \mathrm{softmax}(\vec{h}^{(l)} \Theta^{(l)}),\ \forall l = 0,\dots, L\\
\nonumber  \h{(l)}  &= \sigma(\W{l}{} \h{(l-1)}),\ \forall l = 1,\dots, L\\
\nonumber   \vec{h}^{(0)}  &= \vec{x}
\end{align}
And the update rule:
\begin{align}
W^{(l)}_{t+1}  \leftarrow W^{(l)}_{t} - \eta \sum_{j=l}^L \alpha^{(j)} \nabla _{W^{(l)}} \loss(\f{(j)}, y_t) \label{eqn:odl_update_rule}
\end{align}
Then for each layer we get the following updates (Fig.~\ref{fig:odl_backprop} shows the ODL computational graph):
\begin{align}
    \W{3}{t+1} &= \W{3}{t} -\eta \nabla_{\W{3}{t}} \alpha^{(3)} \loss_3 = \W{3}{t} -\eta \alpha_3 \frac{\partial \loss_3}{\partial \f{(3)}}\frac{\partial \f{(3)}}{\partial \h{(3)}}\frac{\partial \h{(3)}}{\partial \W{3}{t}}\\
    \W{2}{t+1} &= \W{2}{t} -\eta \left(\nabla_{\W{2}{t}} \alpha^{(2)} \loss_2 + \nabla_{\W{2}{t}}  \alpha^{(3)} \loss_3 \right) \nonumber \\
    &= \W{2}{t} -\eta \left(\alpha_2 \frac{\partial \loss_2}{\partial \f{(2)}}\frac{\partial \f{(2)}}{\partial \h{(2)}}\frac{\partial \h{(2)}}{\partial \W{2}{t}} + \alpha_3 \frac{\partial \loss_3}{\partial \f{(3)}}\frac{\partial \f{(3)}}{\partial \h{(3)}}\frac{\partial \h{(3)}}{\partial \h{(2)}}\frac{\partial \h{(2)}}{\partial \W{2}{t}}\right) \\
    \W{1}{t+1} &= \W{1}{t} -\eta \left(\nabla_{\mathbf{W}^{(1)}_t} \alpha^{(1)} \loss_1 + \nabla_{\W{1}{t}} \alpha^{(2)} \loss_2 + \nabla_{\W{1}{t}}  \alpha^{(3)} \loss_3 \right) \nonumber \\
    &= \W{1}{t} -\eta \Bigg(\alpha_1 \frac{\partial \loss_1}{\partial \f{(1)}}\frac{\partial \f{(1)}}{\partial \h{(1)}}\frac{\partial \h{(1)}}{\partial \W{1}{t}} 
    + \alpha_2 \frac{\partial \loss_2}{\partial \f{(2)}}\frac{\partial \f{(2)}}{\partial \h{(2)}}\frac{\partial \h{(2)}}{\partial \h{(1)}}\frac{\partial \h{(1)}}{\partial \W{1}{t}} 
    + \alpha_3 \frac{\partial \loss_3}{\partial \f{(3)}}\frac{\partial \f{(3)}}{\partial \h{(3)}}\frac{\partial \h{(3)}}{\partial \h{(2)}}\frac{\partial \h{(2)}}{\partial \h{(1)}}\frac{\partial \h{(1)}}{\partial \W{1}{t}} \Bigg) 
\end{align}

\begin{figure}
    \centering
    \includegraphics[height=10cm]{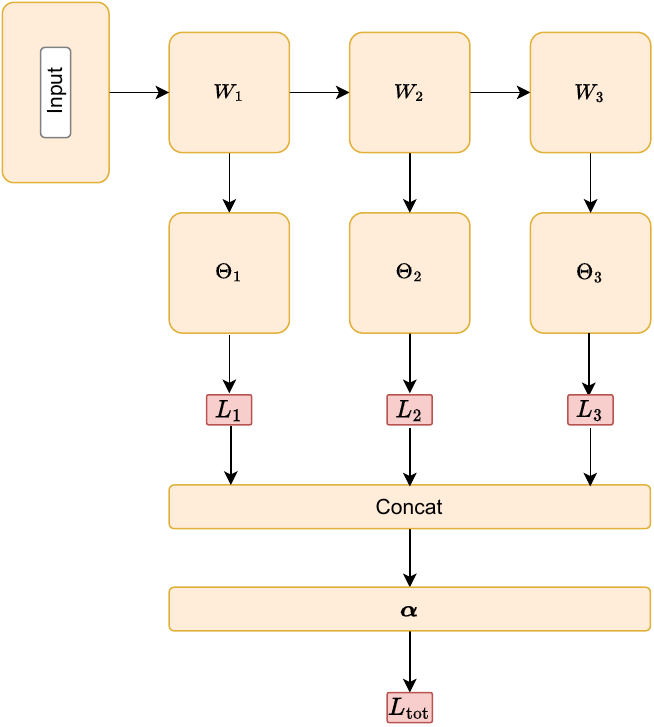}
    \includegraphics[height=9cm]{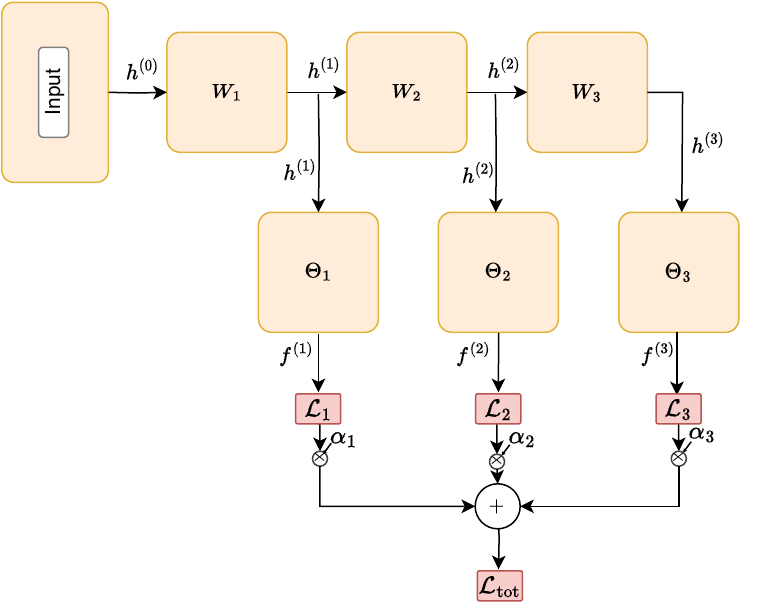}
    \caption{Our proposed architectural modification that provides equivalent total gradients but a more tractable computational graph. There is only one gradient calculation starting from $\loss_{\text{tot}}$. Top and bottom figure show the same architecture, bottom emphasizes that $\alpha_i$ is a constant that is not optimized during backprop.}
    \label{fig:our_backprop}
\end{figure}

Now, we consider our proposed architecture, shown in Fig.~\ref{fig:our_backprop}. In our case the weights can be updated using standard backpropagation, i.e., $\W{l}{t+1} \gets \W{l}{t} - \eta \nabla_{\W{l}{t}} \loss$, where $\loss=\sum_{i=1}^L \alpha^{(i)} \loss_i$.

\noindent \textbf{Lemma 1:} The gradient $\frac{\partial \loss_l}{\partial \W{k}{}} = 0$ for all $(k, l) \in \{1,2,\dots,L\}^2$ with $k > l$.

\emph{Proof:} Suppose $k>l$ and recall from the definition, $\f{(l)} = \mathrm{softmax}(\vec{h}^{(l)} \Theta^{(l)}),$ with 
$\h{(l)} = \sigma(\W{l}{} \h{(l-1)}).$ Now following the recursive definition of $\h{(l)}$ we have:
\begin{align}
    \h{(l)} &= \sigma\left(\W{l}{} \h{(l-1)}\right) = \sigma\left(\W{l}{} \sigma\left(\W{l-1}{} \sigma \left(\cdots\left(\sigma\left(\W{1}{} \h{(0)}\right)\right)\cdots\right)\right)\right) \\
    \f{(l)} &= \mathrm{softmax}(\vec{h}^{(l)} \Theta^{(l)}) = \mathrm{softmax}\left[\sigma\left(\W{l}{} \sigma\left(\W{l-1}{} \sigma \left(\cdots\left(\sigma\left(\W{1}{} \h{(0)}\right)\right)\cdots\right)\right)\right) \boldsymbol{\Theta}^{(l)}\right] \label{eq:lemma}
\end{align}
Observe that for $k>l$, $\W{k}{}$ does not appear in~\cref{eq:lemma} thus we immediately conclude $\frac{\partial \f{(l)}}{\partial \W{k}{}} = 0$. Recall that the operator $\loss_l$ is a function of $\f{(l)}$, \emph{i.e.}, $\loss_l(\f{(l)}, y^{(t)})$.
Given that $y^{(t)}$ is a fixed constant, by the chain rule, $\frac{\partial \loss_l}{\partial \W{k}{}} = \frac{\partial \loss_l}{\partial \f{(l)}}\frac{\partial \f{(l)}}{\partial \W{k}{}}=0$.

\noindent \textbf{Corollary: 1} $\nabla_{\W{k}{t}} \loss = \nabla_{\W{k}{t}} \sum_{i=1}^L \alpha^{(i)} \loss_i = \nabla_{\W{k}{t}} \sum_{i=k}^L \alpha^{(i)} \loss_i$.

\emph{Proof:} All $i<k$ the terms in the sum are equal to zero by direct application of the previous lemma:
\begin{align}
    \nabla_{\W{k}{t}} \loss &= \nabla_{\W{k}{t}} \sum_{i=1}^L \alpha^{(i)} \loss_i = \nabla_{\W{k}{t}} \sum_{i=1}^{k-1} \alpha^{(i)} \loss_i + \nabla_{\W{k}{t}} \sum_{i=k}^L \alpha^{(i)} \loss_i \nonumber \\ 
    &= \underbrace{\nabla_{\W{k}{t}} \alpha^{(1)} \loss_1}_{=0 \text{ by lemma 1}} + \dots + \underbrace{\nabla_{\W{k}{t}} \alpha^{(k-1)} \loss_{k-1}}_{=0 \text{ by lemma 1}} + \nabla_{\W{k}{t}} \sum_{i=k}^L \alpha^{(i)} \loss_i \nonumber \\
    &= \nabla_{\W{k}{t}} \sum_{i=k}^L \alpha^{(i)} \loss_i.
\end{align}
Now, using Lemma 1 and by linearity of differentiation we have:
\begin{align}
    \W{3}{t+1} &= \W{3}{t} - \eta \nabla_{\W{3}{t}} \loss = \W{3}{t} - \eta \nabla_{\W{3}{t}} \sum_{i=1}^3 \alpha^{(i)} \loss_i = \W{3}{t} -\eta \nabla_{\W{3}{t}} \alpha^{(3)} \loss_3 \nonumber\label{eq:w3}\\
    &= \W{3}{t} -\eta \alpha_3 \frac{\partial \loss_3}{\partial \f{(3)}}\frac{\partial \f{(3)}}{\partial \h{(3)}}\frac{\partial \h{(3)}}{\partial \W{3}{t}} \\
    \W{2}{t+1} &= \W{2}{t} - \eta \nabla_{\W{2}{t}} \loss = \W{2}{t} - \eta \nabla_{\W{2}{t}} \sum_{i=1}^3 \alpha^{(i)} \loss_i = \W{2}{t} -\eta \nabla_{\W{2}{t}} \left(\alpha^{(2)} \loss_2 + \alpha^{(3)} \loss_3 \right)\nonumber\\
    &=\W{2}{t} -\eta \left(\alpha_2 \frac{\partial \loss_2}{\partial \f{(2)}}\frac{\partial \f{(2)}}{\partial \h{(2)}}\frac{\partial \h{(2)}}{\partial \W{2}{t}} + \alpha_3 \frac{\partial \loss_3}{\partial \f{(3)}}\frac{\partial \f{(3)}}{\partial \h{(3)}}\frac{\partial \h{(3)}}{\partial \h{(2)}}\frac{\partial \h{(2)}}{\partial \W{2}{t}}\right)\label{eq:w2} \\
    \W{1}{t+1} &= \W{1}{t} - \eta \nabla_{\W{1}{t}} L = \W{1}{t} - \eta \nabla_{\W{1}{t}} \sum_{i=1}^3 \alpha^{(i)} \loss_i = \W{1}{t} -\eta \nabla_{\W{1}{t}} \left(\alpha^{(1)} \loss_1 + \alpha^{(2)} \loss_2 + \alpha^{(3)} \loss_3 \right)\nonumber\\
    &=\W{1}{t} -\eta \Bigg(\alpha_1 \frac{\partial \loss_1}{\partial \f{(1)}}\frac{\partial \f{(1)}}{\partial \h{(1)}}\frac{\partial \h{(1)}}{\partial \W{1}{t}} 
    + \alpha_2 \frac{\partial \loss_2}{\partial \f{(2)}}\frac{\partial \f{(2)}}{\partial \h{(2)}}\frac{\partial \h{(2)}}{\partial \h{(1)}}\frac{\partial \h{(1)}}{\partial \W{1}{t}} \label{eq:w1}
    + \alpha_3 \frac{\partial \loss_3}{\partial \f{(3)}}\frac{\partial \f{(3)}}{\partial \h{(3)}}\frac{\partial \h{(3)}}{\partial \h{(2)}}\frac{\partial \h{(2)}}{\partial \h{(1)}}\frac{\partial \h{(1)}}{\partial \mathbf{W}_t^{(1)}} \Bigg) 
\end{align}

The above 3 equations show that for $L = 3$ and $l \in \{1, 2, 3\}$:
\begin{align}
\mathbf{W}^{(l)}_{t} - \eta \sum_{j=l}^{L} \alpha^{(j)} \nabla _{\mathbf{W}^{(l)}} \loss(\f{(j)}, y_t)
 = \mathbf{W}^{(l)}_{t} - \eta \nabla _{\mathbf{W}^{(l)}}\sum_{j=1}^{L} \alpha^{(j)} \loss(\f{(j)}, y_t)\,.
\end{align}

\textbf{INDUCTIVE HYPOTHESIS: Assume that the proposition holds for} $L=K$, $l \in \{1, \dots, L\}$.
\begin{align}
\mathbf{W}^{(l)}_{t} - \eta \sum_{j=l}^{K} \alpha^{(j)} \nabla _{\mathbf{W}^{(l)}} \loss(\f{(j)}, y_t)
 = \mathbf{W}^{(l)}_{t} - \eta \nabla _{\mathbf{W}^{(l)}}\sum_{j=1}^{K} \alpha^{(j)} \loss(\f{(j)}, y_t)\,.
\end{align}

\textbf{INDUCTIVE STEP: Prove that the proposition holds for $L=K+1$, $l \in \{1, \dots, L\}$}.

We want to show
\begin{align}
\mathbf{W}^{(l)}_{t} - \eta \sum_{j=l}^{K+1} \alpha^{(j)} \nabla _{\mathbf{W}^{(l)}} \loss(\f{(j)}, y_t)
 = \mathbf{W}^{(l)}_{t} - \eta \nabla _{\mathbf{W}^{(l)}}\sum_{j=1}^{K+1} \alpha^{(j)} \loss(\f{(j)}, y_t)\,.
\end{align}

Consider the left hand side of the above equation. And apply inductive hypothesis on first $K$ terms of sum.
\begin{align}
\mathbf{W}^{(l)}_{t} - \eta \sum_{j=l}^{K+1} \alpha^{(j)} \nabla _{\mathbf{W}^{(l)}} \loss(\f{(j)}, y_t)
 &= \mathbf{W}^{(l)}_{t} - \eta \left[ \sum_{j=l}^{K} \alpha^{(j)} \nabla _{\mathbf{W}^{(l)}} \loss(\f{(j)}, y_t)\right] - \eta \alpha^{(l)}\nabla _{\mathbf{W}^{(l)}} \loss(\f{(K+1)}, y_t)\\
 &= \mathbf{W}^{(l)}_{t} - \eta \nabla _{\mathbf{W}^{(l)}}\sum_{j=1}^{K} \alpha^{(j)} \loss(\f{(j)}, y_t) - \eta \alpha^{(K+1)}\nabla _{\mathbf{W}^{(l)}} \loss(\f{(K+1)}, y_t)\\
 &= \mathbf{W}^{(l)}_{t} - \eta \nabla _{\mathbf{W}^{(l)}}\sum_{j=1}^{K+1} \alpha^{(j)} \loss(\f{(j)}, y_t)\,.
\end{align}
This closes the induction step and completes the proof.

\subsection{Complexity analysis}
In this section we compare ODL and our proposed modified optimization setup, which we call Fast-ODL (FODL), in terms of training time complexity.

\noindent \textbf{ODL training complexity:} Consider the computations required to perform an update on hidden layer parameters $\vec{W}^{(l)}$. Based on the update rule~\cref{eqn:odl_update_rule}, to update weight matrix  $\vec{W}^{(l)}$ we need to calculate $L-l+1$ terms in the sum.
Assuming that calculating each partial derivative in the previous expressions takes constant time, and noting that the weight update needs to be done for each layer $l\in\{0,1,\dots,L\}$ we obtain a complexity of:
\begin{align}
    \sum_{j=1}^{L} \sum_{i=j}^L O(1) = \frac{L(L+1)}{2} = O(L^2)
\end{align}
per training step. 
Given that we are in an online learning setting with batch size of 1, in a dataset of size $n$ we are thus going to take $n$ training steps (one step per training datum) for a total training complexity of:
\begin{align}
    \sum_1^n \sum_{j=1}^{L} \sum_{i=j}^L O(1) = O(nL^2)
\end{align}
The quadratic complexity in the number of layers occurs because of redundant calculations. For example, to update both $\vec{W}^{(l)}$ and $\vec{W}^{(l-1)}$ we need to compute $\frac{\partial \loss_l}{\partial \vec{f}^{(l)}}$. However, in the vanilla ODL setup this result is not cached and needs to be recomputed two separate times.

\noindent \textbf{FODL training complexity:}
Looking at~\cref{eqn:odl_update_rule}, we note that the number of gradient calculations would decrease significantly if the the gradient could be taken outside the sum:
\begin{align}
W^{(l)}_{t+1}  \leftarrow W^{(l)}_{t} - \eta  \nabla _{W^{(l)}}\sum_{j=l}^L \alpha^{(j)} \loss(\f{(j)}, y_t). \label{eqn:fodl_update_rule}
\end{align}
However, this expression cannot be directly computed using automatic differentiation as there is no node equal to $\sum_{j=l}^L \alpha^{(j)} \loss(\f{(j)}, y_t)$ is the computational graph in Fig~\ref{fig:odl_backprop}.
By introducing a concatenation and weighted summation of the intermediate losses, as shown in Fig.~\ref{fig:our_backprop}, we can generate this node and backpropagate from it. Then using PyTorch's automatic differentiation framework, computed gradients are cached such that there is no redundant calculation, \emph{i.e.}, when updating $\vec{W}^{(1)}$ we compute $\frac{\partial \loss_l}{\partial \vec{f}^{(l)}}$ and store it. Then for all subsequent calculations for updating $\vec{W}^{(l)}$ we access the cache at negligible computational cost.
This yields a total training cost of:
\begin{align}
    \sum_1^n \sum_{j=1}^{L} nO(1) = O(nL).
\end{align}
For example, consider computing $\vec{W}_{t+1}^{(3)}$ after having computed $\vec{W}_{t+1}^{(1)}$. Recalling the equations for their updates shown earlier in this section, we know that:
\begin{align}
    \W{3}{t+1} &= \W{3}{t} -\eta \alpha_3\mathcolor{teal}{\frac{\partial \loss_3}{\partial \f{(3)}}\frac{\partial \f{(3)}}{\partial \h{(3)}}}\frac{\partial \h{(3)}}{\partial \W{3}{t}} \nonumber \\
    \W{2}{t+1} &= \W{2}{t} -\eta \left(\alpha_2 \mathcolor{orange}{\frac{\partial \loss_2}{\partial \f{(2)}}\frac{\partial \f{(2)}}{\partial \h{(2)}}}\frac{\partial \h{(2)}}{\partial \W{2}{t}} + \alpha_3 \mathcolor{teal}{\frac{\partial \loss_3}{\partial \f{(3)}}\frac{\partial \f{(3)}}{\partial \h{(3)}}}\mathcolor{orange}{\frac{\partial \h{(3)}}{\partial \h{(2)}}}\frac{\partial \h{(2)}}{\partial \W{2}{t}}\right) \nonumber\\
    \W{1}{t+1} &= \W{1}{t} -\eta \Bigg(\alpha_1 \frac{\partial \loss_1}{\partial \f{(1)}}\frac{\partial \f{(1)}}{\partial \h{(1)}}\frac{\partial \h{(1)}}{\partial \W{1}{t}} 
    + \alpha_2 \mathcolor{orange}{\frac{\partial \loss_2}{\partial \f{(2)}}\frac{\partial \f{(2)}}{\partial \h{(2)}}}\frac{\partial \h{(2)}}{\partial \h{(1)}}\frac{\partial \h{(1)}}{\partial \W{1}{t}} \nonumber \\
    &\quad\quad\quad\quad\quad\quad + \alpha_3 \mathcolor{teal}{\frac{\partial \loss_3}{\partial \f{(3)}}\frac{\partial \f{(3)}}{\partial \h{(3)}}}\mathcolor{orange}{\frac{\partial \h{(3)}}{\partial \h{(2)}}}\frac{\partial \h{(2)}}{\partial \h{(1)}} \frac{\partial \h{(1)}}{\partial \W{1}{t}} \Bigg) \nonumber
\end{align}
Clearly, caching the term in \textcolor{teal}{teal} would reduce computation. Thus, in FODL computing the update for the first layer $\W{1}{t+1}$ and storing all the partial derivatives along the computational graph provides all the necessary information to calculate $\W{2}{t+1}, \dots, \W{L}{t+1}$ with computing only one new derivative $\frac{\partial \vec{h}^{(l)}}{\partial\W{l}{t}}$. Similarly this holds for computing $\W{2}{t+1}$ after $\W{1}{t+1}$, see \textcolor{orange}{orange} terms. 

\noindent More precisely, in the standard ODL formulation the redundancy of the \textcolor{teal}{teal} term is very high as it is computed $L$ times  (the terms in \textcolor{orange}{orange} $L-1$ times) but in our formulation every gradient is only computed once. This yields a total computational saving of:
\begin{align}
    \sum_{1}^n\sum_{j=1}^L\sum_{i=j+1}^L=n\frac{L(L-1)}{2}O(1).
\end{align}
By subtracting the amount of computation saved by FODL from the total complexity of the base ODL method we obtain the complexity of our proposed approach and it is linear:
\begin{align}
    \sum_{1}^n\sum_{j=1}^L\sum_{i=j}^L O(1)- \sum_{1}^n\sum_{j=1}^L\sum_{i=j+1}^L O(1) = n\left(\frac{L(L+1)}{2} - \frac{L(L-1)}{2}\right)O(1) = nLO(1) = O(nL).
\end{align}

\end{document}